\documentclass[12pt,a4paper]{article}

\usepackage[cmex10,intlimits]{amsmath}
\usepackage{wasysym}
\usepackage{amssymb}
\usepackage{amsthm}
\usepackage{amsfonts}
\usepackage{amsmath}
\usepackage{color}
\usepackage{a4wide}
\usepackage{rotating}
\usepackage{cite}

\makeatletter
\def\blfootnote{\xdef\@thefnmark{}\@footnotetext}
\makeatother

\title{On Measuring and Quantifying Performance:\\Error Rates, Surrogate Loss, and an Example in SSL}

\author{
\small\begin{tabular}{r|l}
\multicolumn{2}{c}{\large Marco~Loog} \medskip \\
Pattern Recognition Laboratory & The Image Section \\
Delft University of Technology & University of Copenhagen \\
The Netherlands & Denmark \\
e-mail: m.loog@tudelft.nl & http: prlab.tudelft.nl \medskip\\
\multicolumn{2}{c}{\large Jesse~Krijthe} \medskip \\
Department of Molecular Epidemiology & Pattern Recognition Laboratory \\
Leiden University Medical Center & Delft University of Technology \\
The Netherlands & The Netherlands \\
e-mail: jkrijthe@gmail.com & http: www.jessekrijthe.com \medskip \\
\multicolumn{2}{c}{\large Are~C.~Jensen} \medskip \\
& Department of Informatics \\
& University of Oslo \\
& Norway \\
e-mail: arej@ifi.uio.no & http: www.mn.uio.no
\end{tabular}}

\date{February 28, 2016\footnote{Appeared as Chapter 1.2 in C.H. Chen. \emph{Handbook of Pattern Recognition and Computer Vision.} 5th edition, World Scientific, 2015}}

\begin{document}

\maketitle

\begin{abstract}
In various approaches to learning, notably in domain adaptation, active learning, learning under covariate shift, semi-supervised learning, learning with concept drift, and the like, one often wants to compare a baseline classifier to one or more advanced (or at least different) strategies.  In this chapter, we basically argue that if such classifiers, in their respective training phases, optimize a so-called surrogate loss that it may also be valuable to compare the behavior of this loss on the test set, next to the regular classification error rate. It can provide us with an additional view on the classifiers' relative performances that error rates cannot capture.  As an example, limited but convincing empirical results demonstrates that we may be able to find semi-supervised learning strategies that can guarantee performance improvements with increasing numbers of unlabeled data in terms of log-likelihood.  In contrast, the latter may be impossible to guarantee for the classification error rate.
\end{abstract}


%

\newpage

\section{Introduction}

The aim of semi-supervised learning is to improve supervised learners by exploiting potentially large amounts of, typically easier to obtain, unlabeled data \cite{chapelle06b}.  Up to now, however, semi-supervised learners have reported mixed results when it comes to such improvements: it is not always the case that semi-supervision results in lower expected error rates.  On the contrary, severely deteriorated performances have been observed in empirical studies and theory shows that improvement guarantees can often only be provided under rather stringent conditions \cite{castelli95a,ben-david08a,lafferty07a,singh08a}.

Now, the principal suggestion put forward in this chapter is that, when dealing with semi-supervised learning, one may not only want to study the (expected) error rates these classifiers produce, but also to measure the classifiers' performances by means of the intrinsic loss they may be optimizing in the first place.  That is, for classification routines that optimize a so-called surrogate loss at training time---which is what many machine learning and Bayesian decision theoretic approaches do \cite{scholkopf2002learning,robert2001bayesian}, we propose to also investigate how this loss behaves on the test set as this can provide us with an alternative view on the classifier's behavior that a mere error rate cannot capture.

In fact, though the main example is concerned with semi-supervision, we would like to argue that in other learning scenarios, similar considerations might be beneficial.  For instance in active learning \cite{settles2010active}, where rather than sampling randomly from ones input data to provide these instances with labels, one aims to do the sampling in a systematic way, trying to keep labeling cost as low as one can or, similarly, to learn from as few labeled examples as possible.  Also here it may (or, we believe, it should) be of interest to not only compare the error rates that different approaches (e.g. random sampling and uncertainty sampling \cite{lewis1994sequential}) achieve, but also how the surrogate losses compare for these techniques when we are using the same underlying classifiers.  Similar remarks now can be made for other learning scenarios like domain adaptation, transfer learning, and learning under data shift or data drift \cite{margolis2011literature,torrey2009transfer,quionero2009dataset,vzliobaite2010learning}.  In these last settings, one may typically want to compare, say, a classifier trained in the source domain with one that exploits additional knowledge on the target domain.

\subsection{Surrogate Loss vs. Error Rates}

The simple idea underlying the suggestion we make is that, unless we make particular assumptions, generally, we cannot expect to minimize the error rate if we are, in fact, optimizing a surrogate loss.  This surrogate loss is, to a large extent, chosen for computational reasons, but of course the hope is that, with increasing training set size, minimizing it will not only lead to improvements with respect to this surrogate loss but also with respect to the expected error rate.  This cannot be guaranteed in any strict way however.  To start with, the classifier's error rate itself can already act rather unpredictably.  A general result by Devroye demonstrates, for instance, that for any classifier there exists a classification problem such that the error rate converges at an arbitrarily slow rate to the Bayes error \cite{devroye1982any}.  If the classifier is not a universal approximator \cite{devroye1996probabilistic,steinwart05}, there is not even a guarantee that the Bayes error will ever be reached.  Worse even, in the case that we are dealing with such model misspecification, error rates might even go up with increasing numbers of training samples \cite{loog12dip}. This leads to the rather counterintuitive result that, in some cases, expected error rates might actually be improved by throwing arbitrary samples out of the training set.  The aforementioned considerations lead us, all in all, to speculate that any kind of generally valid (i.e., not depending on strong assumptions) expected performance guarantees, if at all possible in semi-supervised learning or any of the other aforementioned learning scenarios, can merely be obtained in terms of the surrogate loss of the classifier at hand.  Overall, these ideas are in line with those presented in \cite{loog15}.

We could definitely imagine that, still, one takes the position that the mere loss that matters is the 0/1 loss and that it is this quantity that has to be minimized.  As far as we can see, however, taking this stance to the extreme, one cannot do anything else than try and directly minimize this 0/1 loss and face all the computational complications that go with it.  On a less philosophical level, one may claim that the 0/1 loss is, in the end, also not the loss that one is interested in.  One might actually have an application-relevant loss and in real applications (clinical, domestic, industrial, pedagogic, etc.) this is but seldom the 0/1 loss.  In fact, the true loss of interest related to a particular classification problem may ultimately be unknown.

For us there is, however, a more basic reason for studying the surrogate loss intrinsic to the classifier at hand. As a matter of a fact, a lower loss really means the model is better, in the sense that the estimated parameters get closer to those of the optimal classifier one would obtain if all data is labeled.  In the particular setting of semi-supervised learning, a decrease in expected loss, when adding unlabeled data, really indicates that the same effect---i.e., an improved model fit---is achieved as with adding more labeled data.  In our opinion, this seems the least we could ask for in a semi-supervised setting.  With this we still do not mean to claim that the surrogate loss is \emph{the} quantity to study, but it does give us a different perspective on the problem of various learning scenarios.  Finally, let us point out that the connection between the 0/1 loss and surrogate losses has in recent years attracted quite some attention.  Some papers investigating theoretical aspects for particular classes of loss functions, but also covering the design of such surrogate losses, are \cite{ben2012minimizing,masnadi2008design,nguyen2009surrogate,reid2009surrogate,reid2010composite,scott2011surrogate}. These contributions follow earlier works such as \cite{bartlett2006convexity}, \cite{buja}, and \cite{zhang2004statistical}.

\subsection{Outline}

This chapter illustrates our point by means of two classifiers that optimize the log-likelihood of the model fit to the data. Clearly, this objective should be maximized, but taking minus the likelihood would turn it into a loss (which is sometimes referred to as the log loss).  The particular classifiers under consideration are the nearest means classifier (NMC) \cite{duda72a} and classical linear discriminant analysis (LDA) \cite{rao1948utilization}.    Next section starts off with a general reflection on these two classifiers after which two semi-supervised variations are introduced.  Section \ref{sect:exp} reports on the results of the experiments, comparing the semi-supervised learners and their supervised counterparts empirically.  The final section discusses our findings in the light of the point we would like to make and concludes this chapter.

\section{A Biased Introduction to Semi-Supervision}

Before we get to semi-supervised NMC and LDA, we feel the need to remark that their regular supervised versions are still capable of providing state-of-the-art performance. Especially for relatively high-dimensional, small sample problems NMC may be a particularly good choice.  Some rather recent examples demonstrating this can be found in bioinformatics and its applications \cite{wilkerson2012differential,villamil2012colon,budczies2012remodeling}, but also in neurology \cite{jolij2011act} and pathology \cite{gazinska2013comparison}. Further use of the NMC can be found in high-impact journals from the fields of oncology, neuroscience, general medicine, pharmacology, and the like.  A handful of the latest examples can be found in \cite{hyde2012mnemonic,haibe2012three,desmet2013identification,sjodahl2012molecular}.  Similar remarks can be made about LDA, though in comparison with the NMC, there should be relatively more data available to make it work at a competitive level.  Like for the NMC, many recent contributions from a large number of disciplines still employ this classical decision rule, e.g.\ \cite{ackermann2013detection,allen2013network,chung2013single,brunton2013rats,price2012cyanophora}.  All in all, like any other classifier, NMC and LDA have their validity and cannot be put aside as being outdated or not-state-of-the-art. The fact that classifiers having been around for 40 years or more, does not mean they are superseded.  In this respect, the reader might also want to consult relevant works such as \cite{handXXX} and \cite{efronXXX}.

\subsection{Supervised NMC and LDA}

The two semi-supervised versions of both the NMC and LDA are those based on classical expectation maximization or self-learning and those based on a so-called intrinsically constrained formulation.  These approaches are introduced in the subsections that follow.  The models underlying supervised NMC and LDA are based on normality assumptions for the class-conditional probability density functions.  More specifically:
\begin{itemize}

\item

LDA is the classical technique where the class-conditional covariance matrices are assumed the same across all classes, but where both the class means and the class priors can vary from class to class.  Estimating these variables under maximum likelihood results in the well-known solutions for the priors and the means, while the overall class covariance matrix becomes the prior weighted sum of the ML estimates of the individual class covariance matrices.

\item

For the NMC the parameter space is further restricted.  In addition to the covariance matrix being the same for all classes it is also constrained to be the a multiple of the identity matrix.  Moreover, the priors are fixed to be equal for all classes.  In \cite{loog15} one can find the solution to this parameter estimation problem.  Here we note that this model is not necessarily unique: there are of course various ways in which one can formulate the NMC (as well as other classifiers) in terms of an optimization problem.  Ours is but one choice.

\end{itemize}

\subsection{EM and Self-Learning}

Self-learning or self-training is a rather generally applicable approach to semi-supervised learning \cite{basu02a,mclachlan75a,vittaut02}.  In an initial step, the classifier of choice is trained on the available labeled data.  Using this trained classifier all unlabeled data is assigned a label.  Then, in a next step all of this now labeled data is added to the training set and the classifier is retrained with this enlarged set.  Given this newly trained classifier one can relabel the initially unlabeled data and retrain the classifier again with these updated labels.  This process is then iterated until convergence, i.e., when the labeling of the initially unlabeled data remains unchanged.  The foregoing only gives the basic recipe for self-learning.  Many variations and alternatives are possible, e.g., one can only take a fraction of the unlabeled data into account when retraining, once labeled one can decide to not relabel the data, etc.

Another well-known, and arguably more principled semi-supervised approach treats the absence of certain labels as a missing data problem. Most of the time this is formulated in terms of a maximum likelihood objective \cite{dempster77a} and relies on the classical technique of expectation maximization (EM) to come to a solution \cite{nigam98a,oneill78a}. Although self-learning and EM may at a first glance seem different ways of tackling the semi-supervised classification problem, \cite{basu02a} effectively shows that self-learners optimize the same objective as EM does (though they may typically end up in different local optima). Similar observations have been made in \cite{abney04a,haffari07a}.

A major problem with EM and self-learning strategies is the fact that they often suffer from severely deteriorated performance with increasing numbers of unlabeled samples. This behavior, which has been extensively studied in various previous works \cite{cohen04a,cozman06a,loog2013semi,yang2011effect}, is typically caused by model misspecification, i.e., the setting in which the statistical model does not fit the actual data distribution.  We note that this is at contrast with the supervised setting, where most classifiers are capable of handling mismatched data assumptions rather well and adding more labeled data typically improves performance.  NMC will most definitely suffer from model misspecification, because of the rather rigid, low-complexity nature of this classifier.  LDA is more flexible, but still only able to model linear decision boundaries.  Hence, also LDA will often be misspecified.

\subsection{Intrinsically Constrained NMC}

In \cite{loog10x} and \cite{loog12a}, a novel way to learn in a semi-supervised manner was introduced.  On a conceptual level, the idea is to exploit constraints that are known to hold for the NMC and LDA and that define relationships between the class-specific parameters of those classifiers and certain statistics that are independent of the particular labeling.  These relationships are automatically fulfilled in the supervised setting but typically impose constraints in the semi-supervised setting.  Specifically, for NMC and LDA the following constraint holds (see \cite{fukunaga90}):
\begin{equation}\label{eq:altlaw}
N m = \sum_{k=1}^K N_k m_k \, ,
\end{equation}
where $K$ is the number of classes, $m$ is the overall sample mean of the data, and $m_k$ are the different sample means of the $K$ classes.  $N$ is the total number of training instances and $N_k$ is the number of observations for class $k$.  For LDA there is an additional constraint that holds (again see \cite{fukunaga90}):
\begin{equation}\label{eq:cov}
 B + W = T \, .
\end{equation}
It relates the standard estimates for the average class-conditional covariance matrix $W$, the between-class covariance matrix $B$, and the estimate of the total covariance matrix $T$.  $W$ is the covariance matrix that models the spread of every class in LDA.

In the supervised setting these constraints do not need to be assumed as they are automatically fulfilled.  Their benefit only becomes apparent with the arrival of unlabeled data. In the semi-supervised setting, the label-independent estimates $m$ and $T$ can be improved.  Using these more accurate estimates, however, results in a violation of the constraints. Fixing the constraints again by properly adjusting $m_i$, $W$, and $B$, these label-dependent estimates become more accurate and in expectation lead to improved classifiers.  For a more detailed account of how to enforce these constraints, we refer to \cite{loog2013semi} (see \cite{krijthe} and \cite{loog} for related approaches).

The constrained estimation approach is less generally applicable, but it can avoid the severe deteriorations self-learning displays: when the model does not match the data, the model fit will obviously not be good, but the constrained semi-supervised fit will generally still be better, in terms of the error rate, than the supervised equivalent.  Still, also in this constrained setting, the results turn out not to be univocal either.  Error rates can increase with increasing number of unlabeled samples and we consider further insight into this issue paramount for a deeper understanding of the semi-supervised learning problem in general.

\section{Experimental Setup and Results}\label{sect:exp}

For the experiments, we used eight data sets from the UCI Machine Learning Repository \cite{Bache+Lichman:2013}, all having two classes. The data sets used, together with some basic specifications, can be found in Table \ref{tab:real}.  We put up the experiments in a way similar to those performed in \cite{loog2013semi}.

\begin{table}[ht]
\begin{center}
\caption{Basic properties of the eight two-class data sets from the UCI Machine Learning Repository \cite{Bache+Lichman:2013}.}\label{tab:real}
\begin{tabular}{l|c|c|c|}
data & \# objects  & dimensions & smallest prior
\\ \hline
{ \tt haberman } & 306 & 3 & 0.26 \\
{ \tt ionosphere } & 351 & 33 & 0.36 \\
{ \tt pima } & 768 & 8 & 0.35 \\
{ \tt sonar } & 208 & 60 & 0.47 \\
{ \tt spect } & 267 & 22 & 0.21 \\
{ \tt spectf } & 267 & 44 & 0.21 \\
{ \tt transfusion } & 748 & 3 & 0.24 \\
{ \tt wdbc } & 569 & 30 & 0.37
\end{tabular}
\end{center}
\end{table}

Experiments with the three NMCs were done for two different total labeled training set sizes, four and ten, while the unlabeled training set sizes considered are $2^1=2$, $2^3$, \dots, $2^{9}$, and $2^{11} = 2048$.  For the supervised and semi-supervised LDAs, experiments were carried out with $100$ labeled samples, while the unlabeled training set sizes are the same as for the NMCs.   In the experiments, we study learning curves for increasing numbers of unlabeled data.  For every combination of the amount of unlabeled objects and labeled objects, 1000 repetitions of randomly drawn data were used to obtain accurate performance estimates.  In order to be able to do so based on the limited amount of samples provided by the data sets, instances were drawn with replacement.  This basically means that we assume that the empirical distribution of every data set is its true distribution and this therefore allows us to measure the true error rates and the true log-likelihoods. It enabled us to properly study our learning curves on real-world data without having to deal with the extra variation due to cross validation and the like.

\begin{figure*}[ht]
\centering
\hrulefill~{\small NMC / error rates / 4 training samples}~\hrulefill \smallskip \\
\includegraphics[width=0.32\hsize]{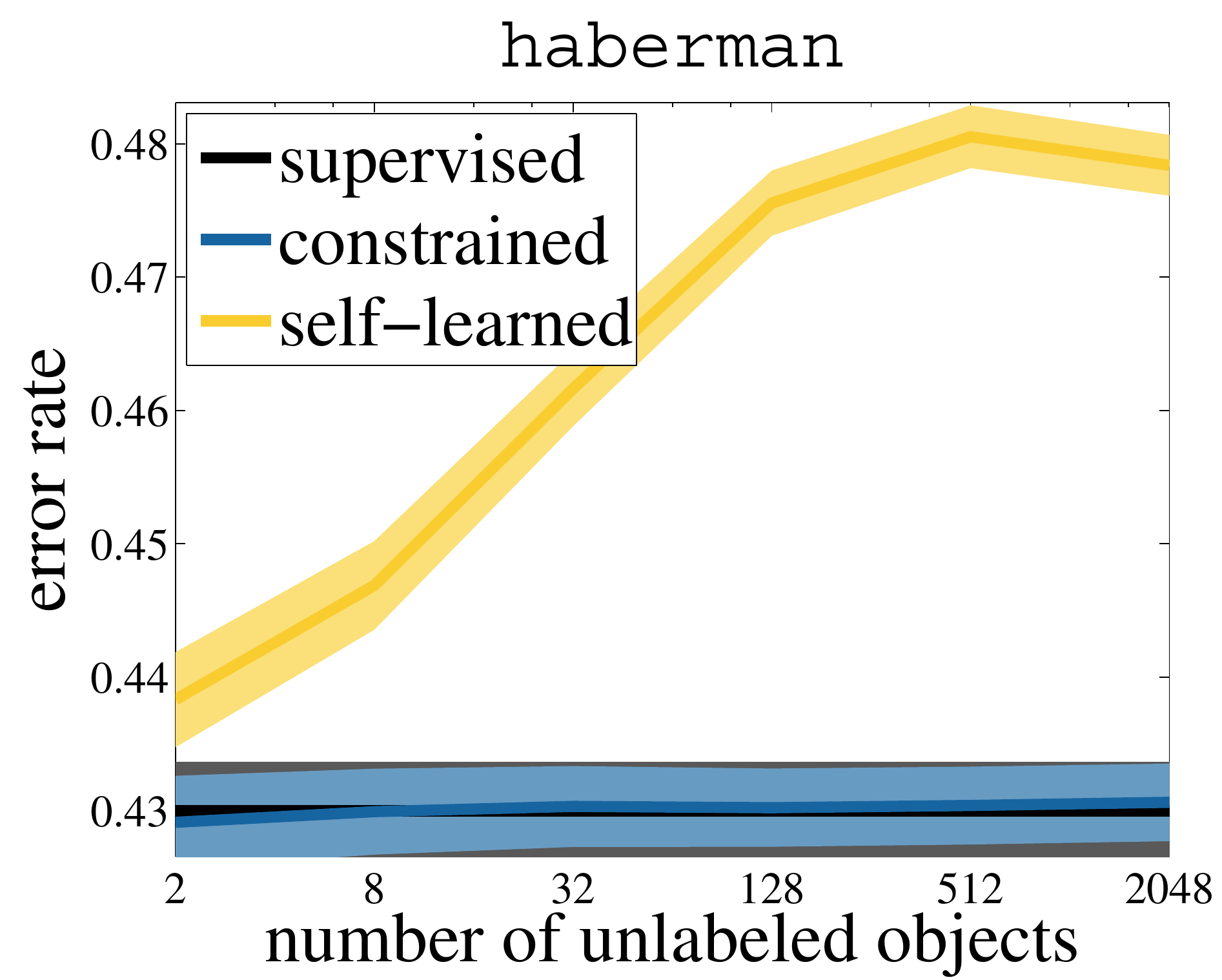}
\includegraphics[width=0.32\hsize]{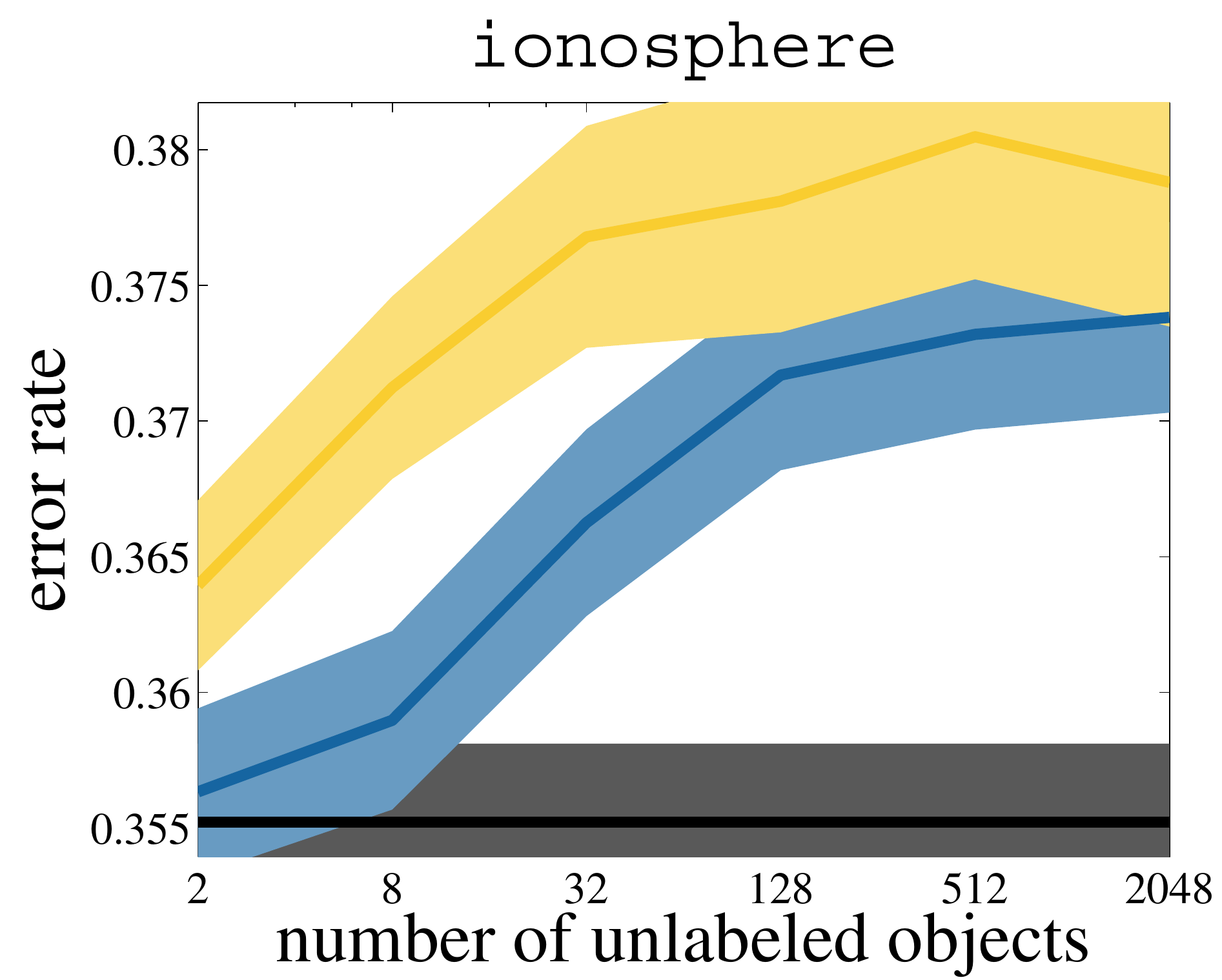}
\includegraphics[width=0.32\hsize]{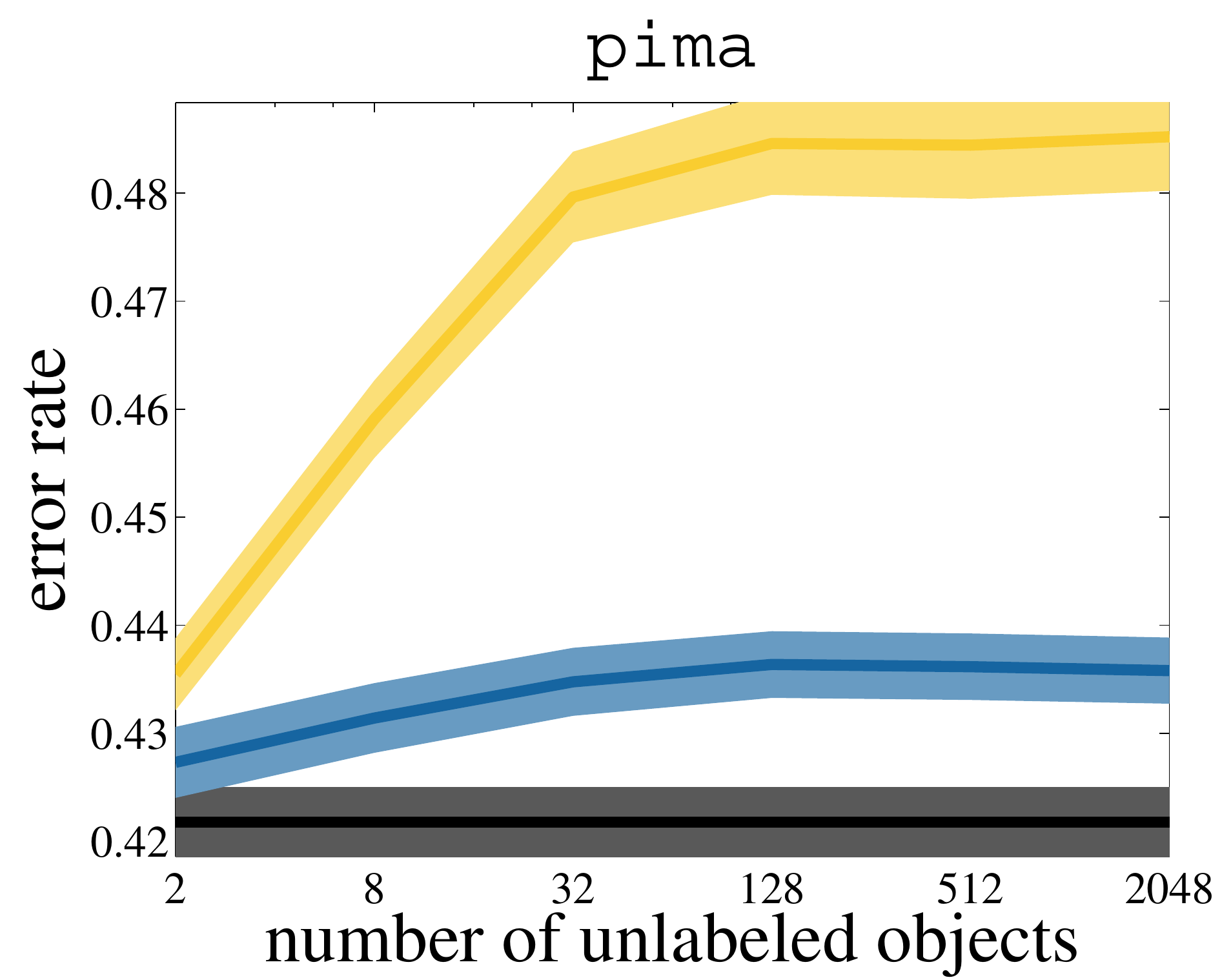} \bigskip \\
\includegraphics[width=0.32\hsize]{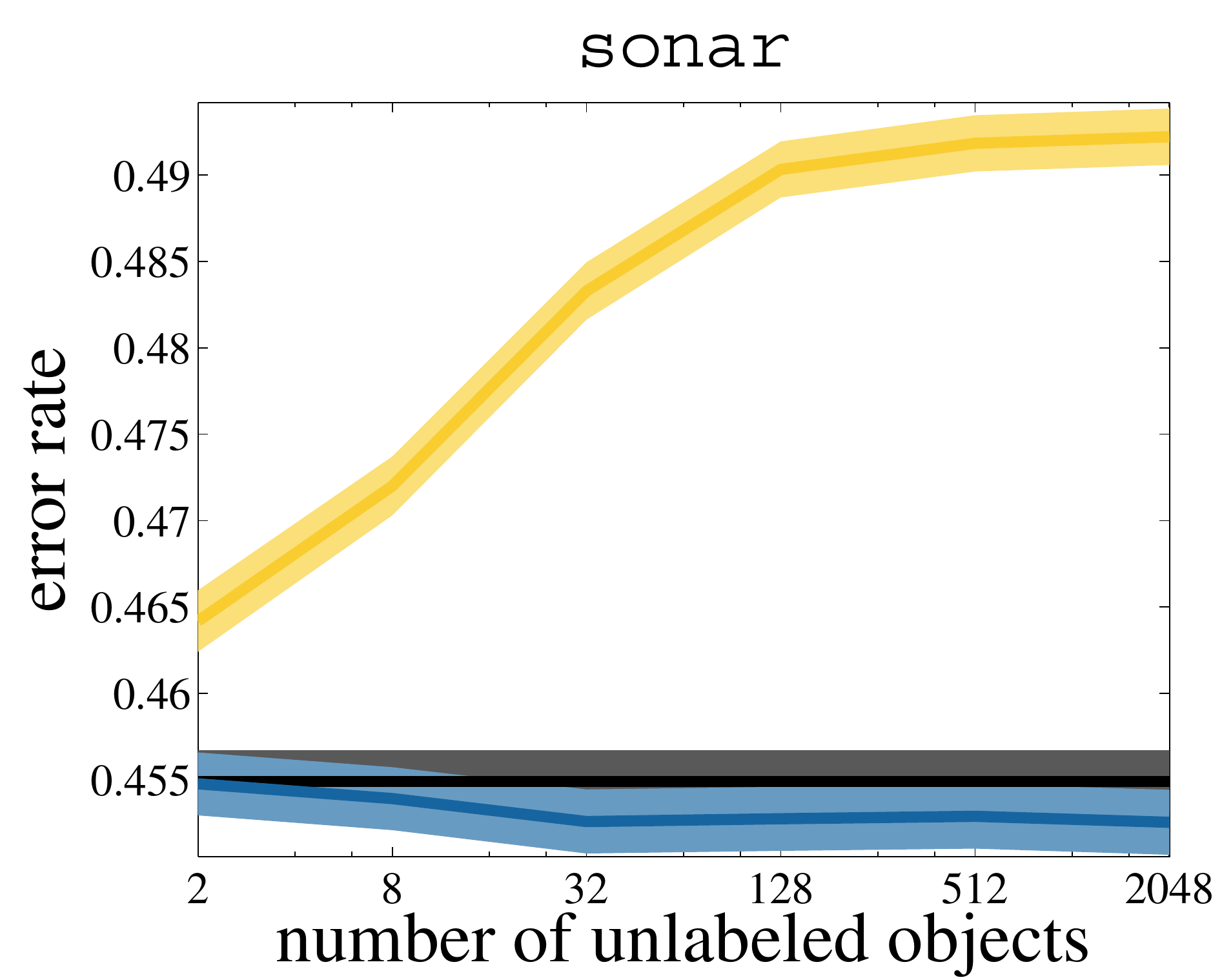}
\includegraphics[width=0.32\hsize]{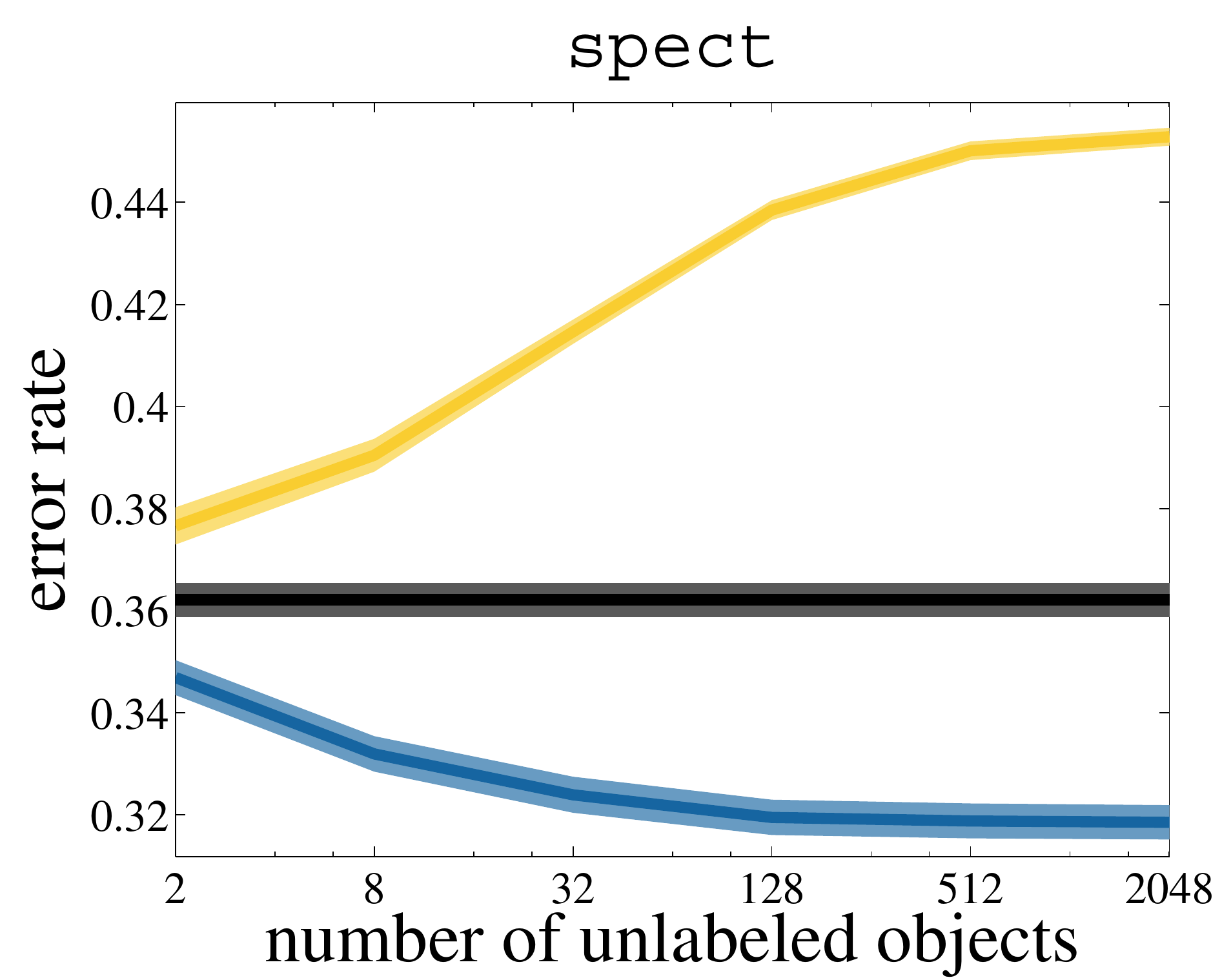} \bigskip \\
\includegraphics[width=0.32\hsize]{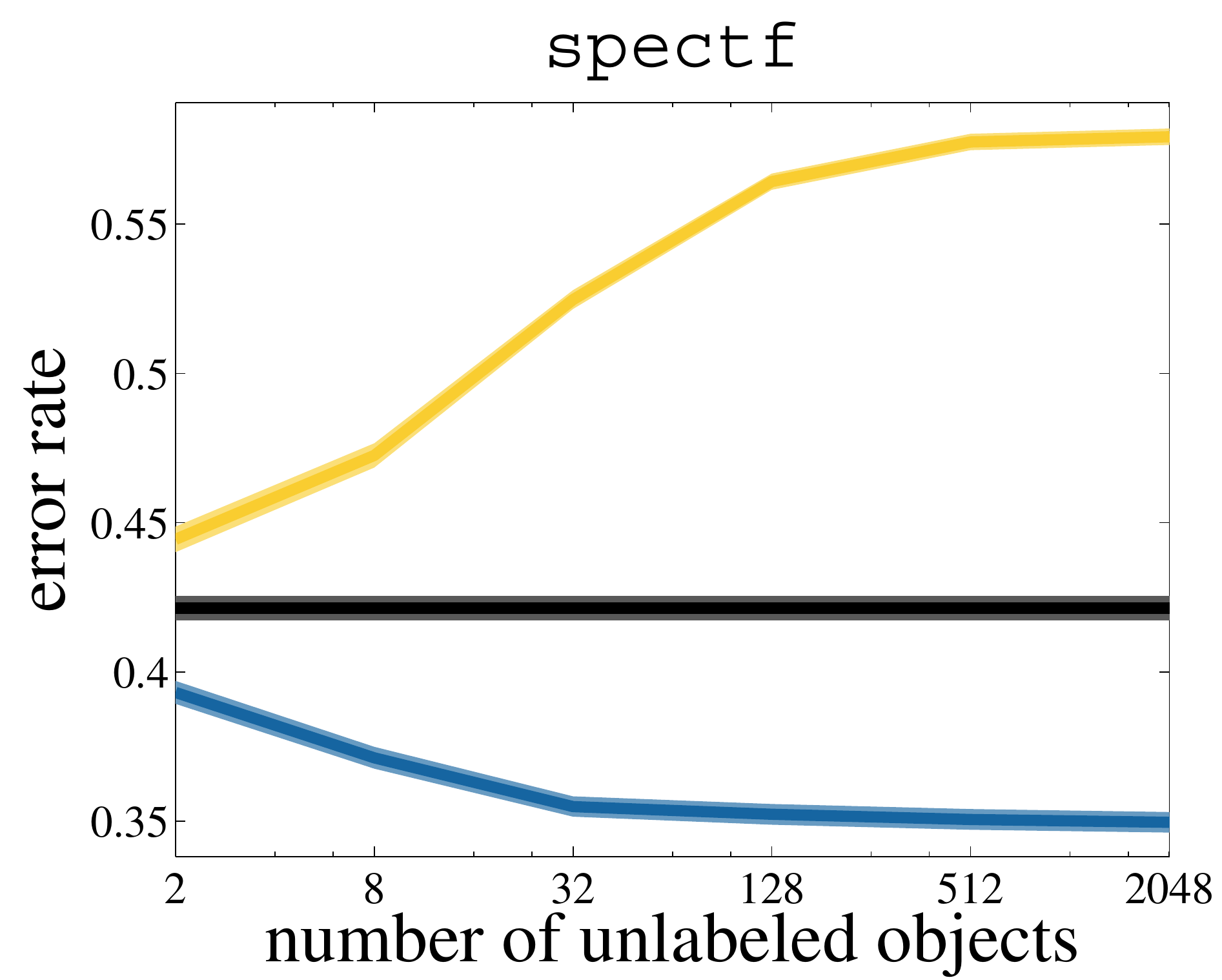}
\includegraphics[width=0.32\hsize]{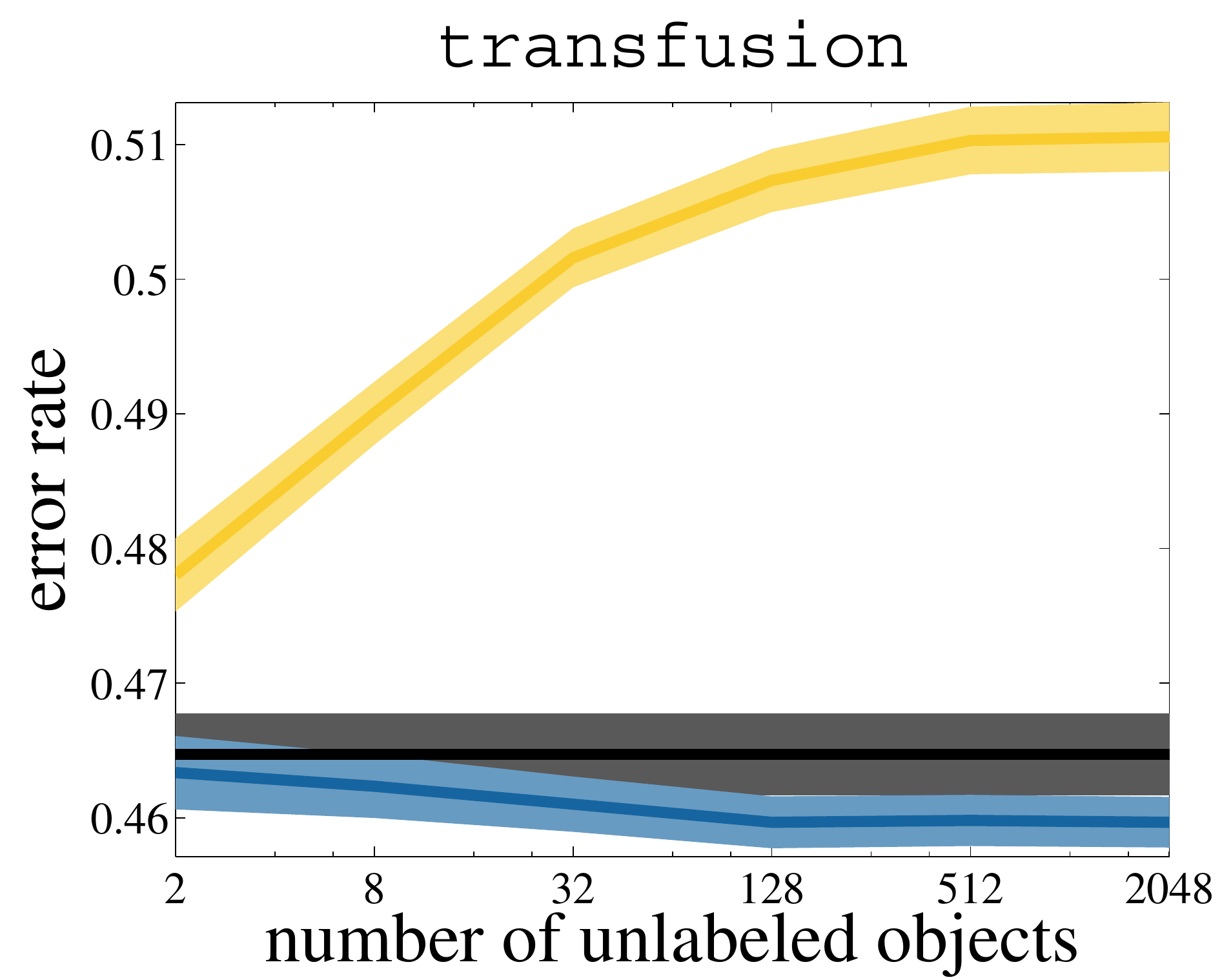}
\includegraphics[width=0.32\hsize]{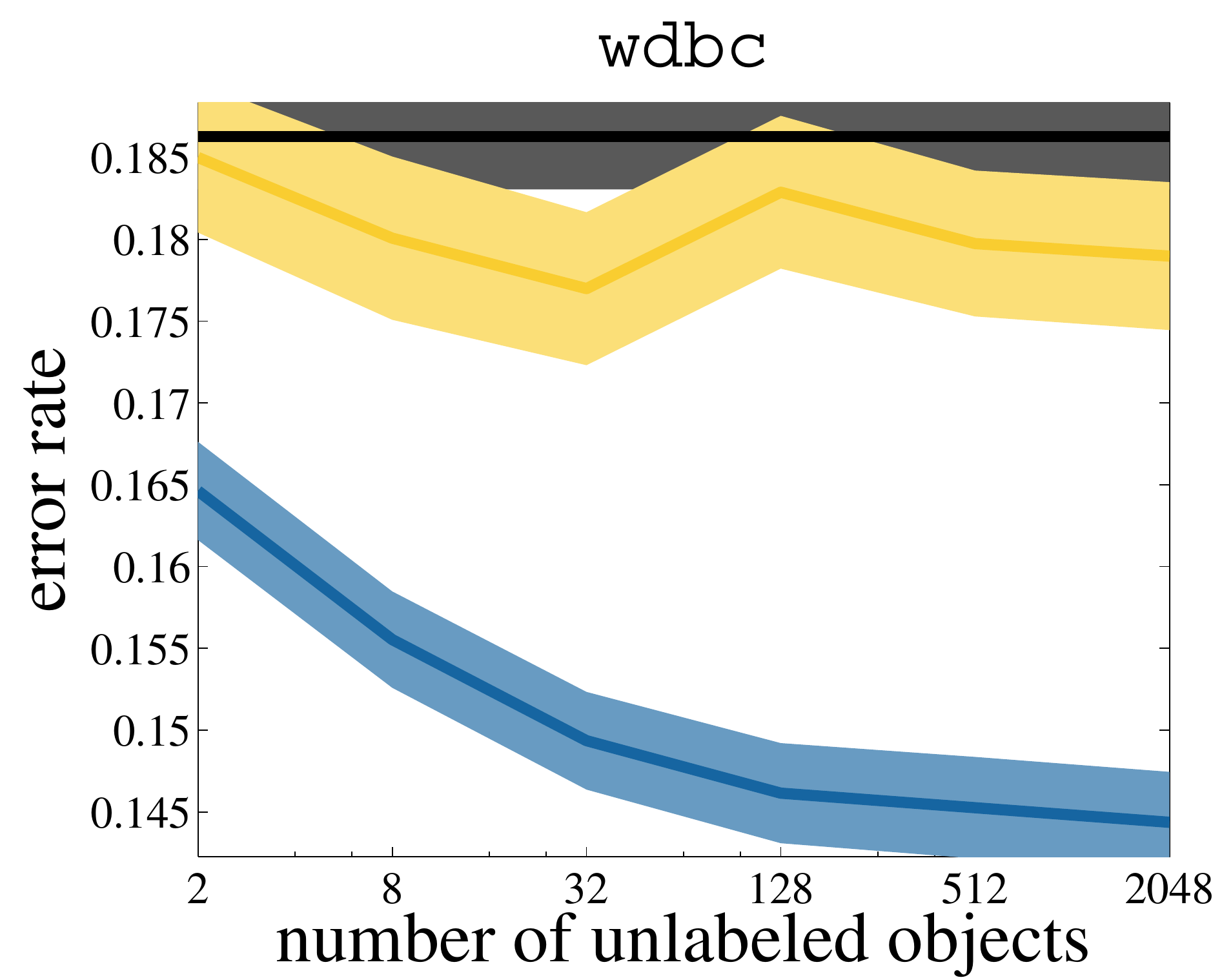}
\caption{Mean error rates for the supervised (black), self-learned (yellow), and the constrained NMC (blue) on the eight real-world data sets for various unlabeled sample sizes and a total of four labeled training samples.}\label{fig:one}
\end{figure*}

\begin{figure*}[ht]
\centering
\hrulefill~{\small NMC / error rates / 10 training samples}~\hrulefill \smallskip \\
\includegraphics[width=0.32\hsize]{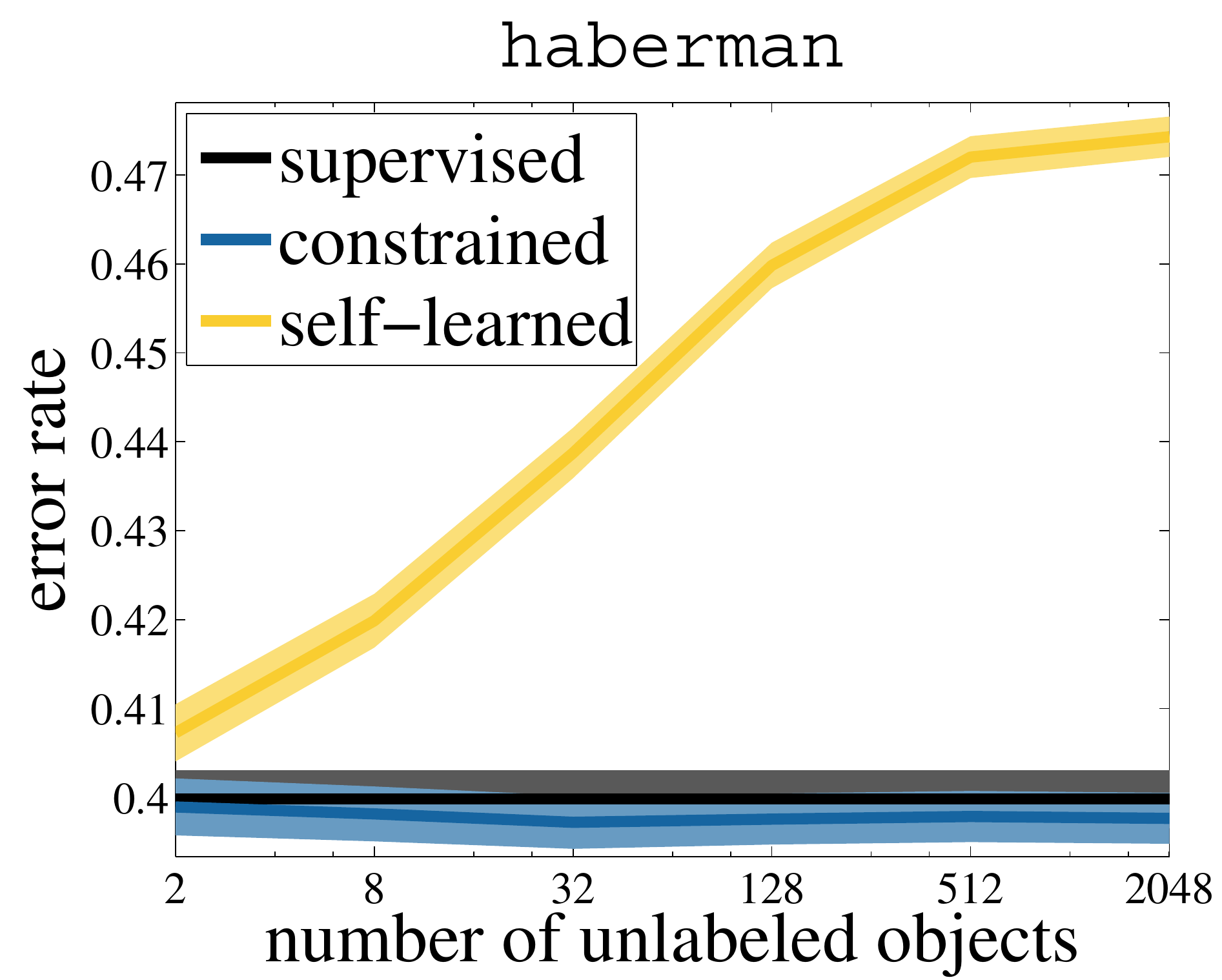}
\includegraphics[width=0.32\hsize]{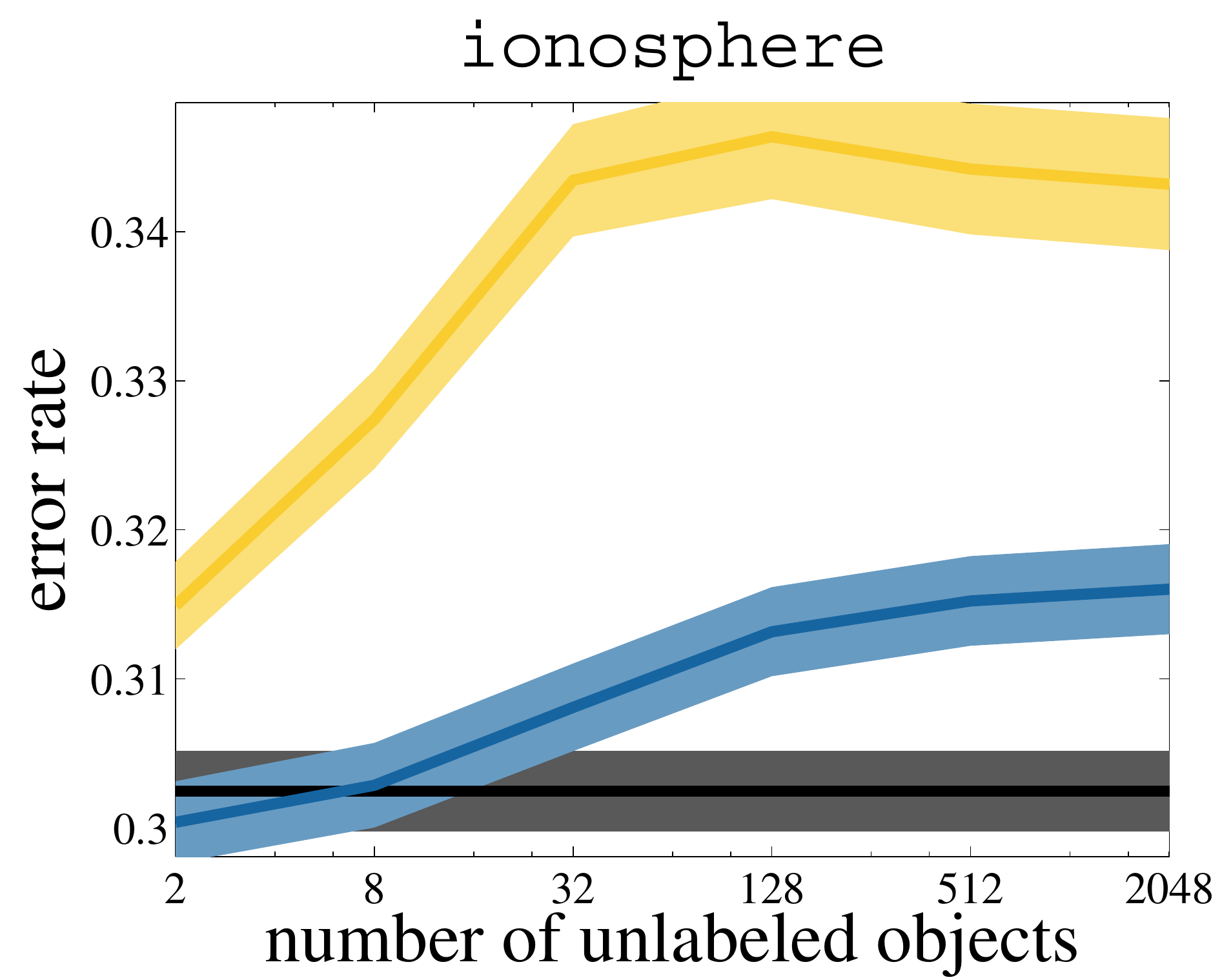}
\includegraphics[width=0.32\hsize]{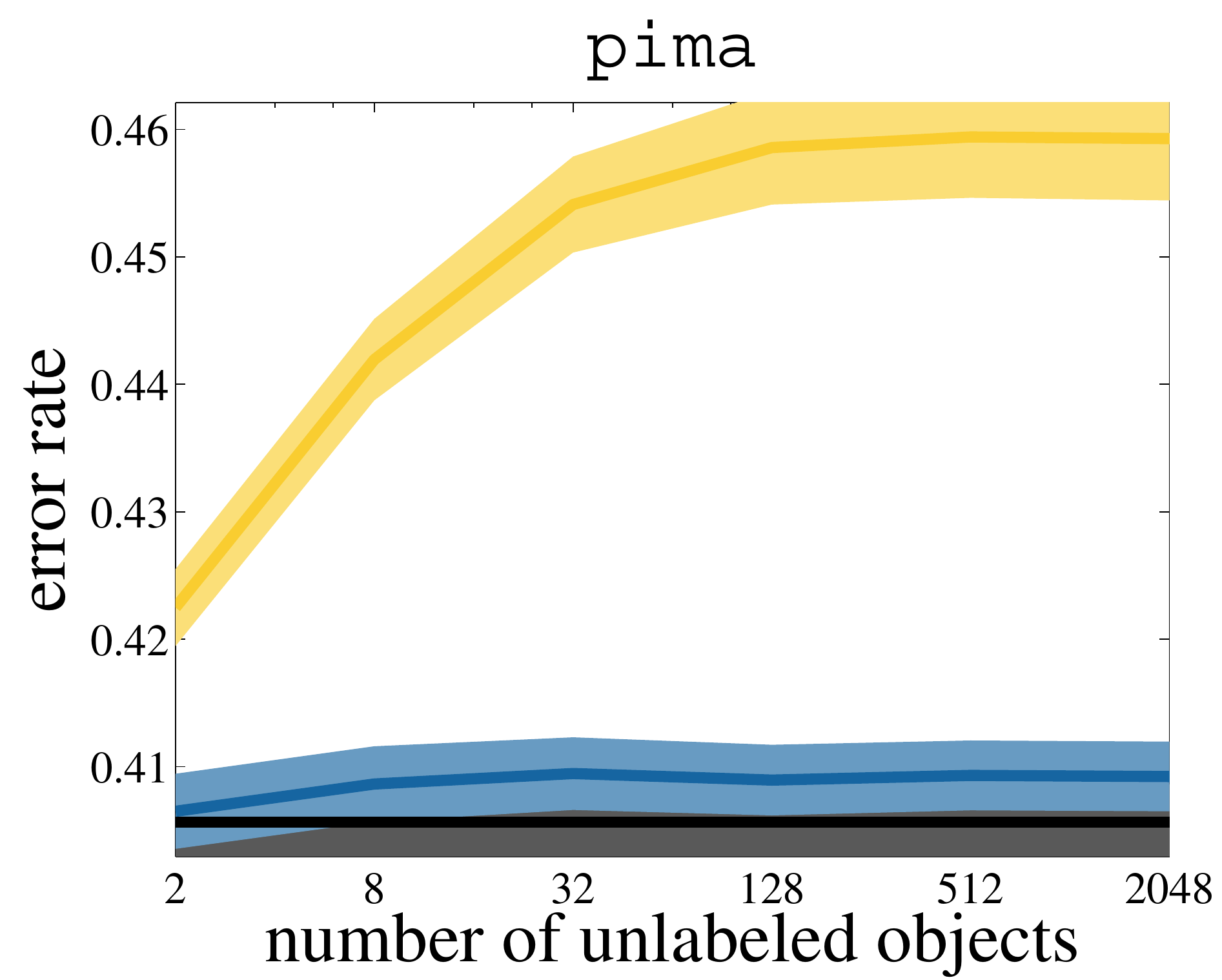} \bigskip \\
\includegraphics[width=0.32\hsize]{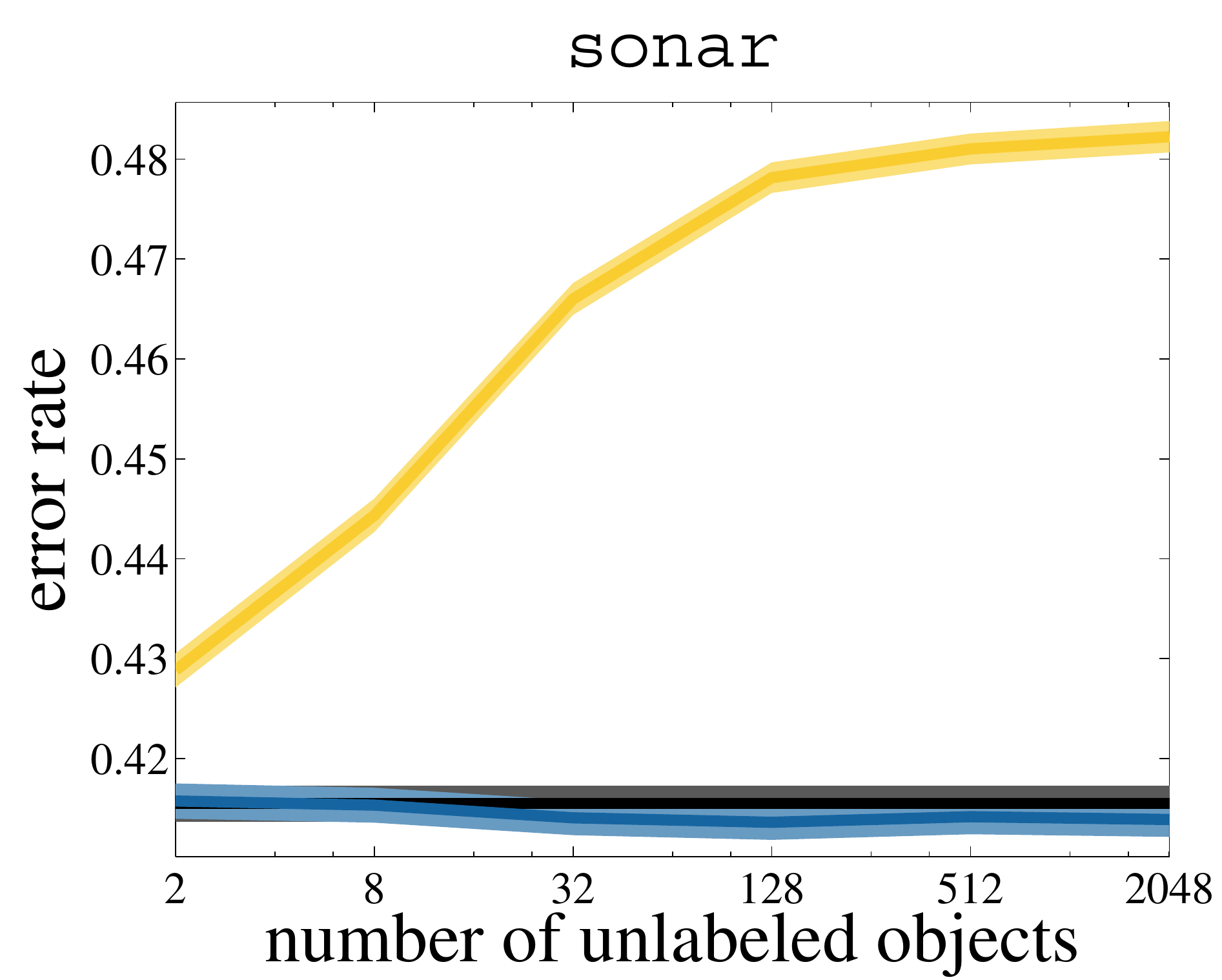}
\includegraphics[width=0.32\hsize]{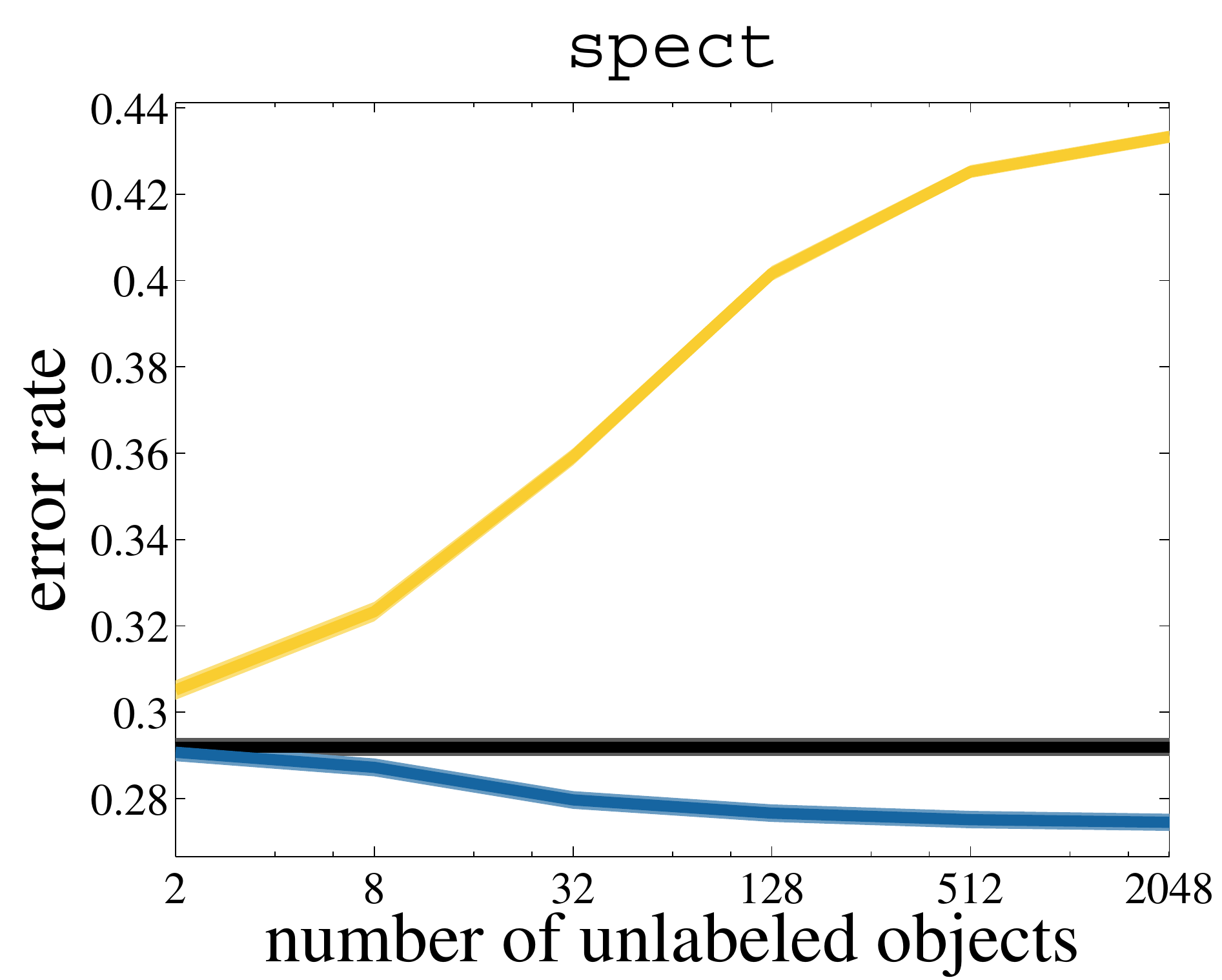} \bigskip \\
\includegraphics[width=0.32\hsize]{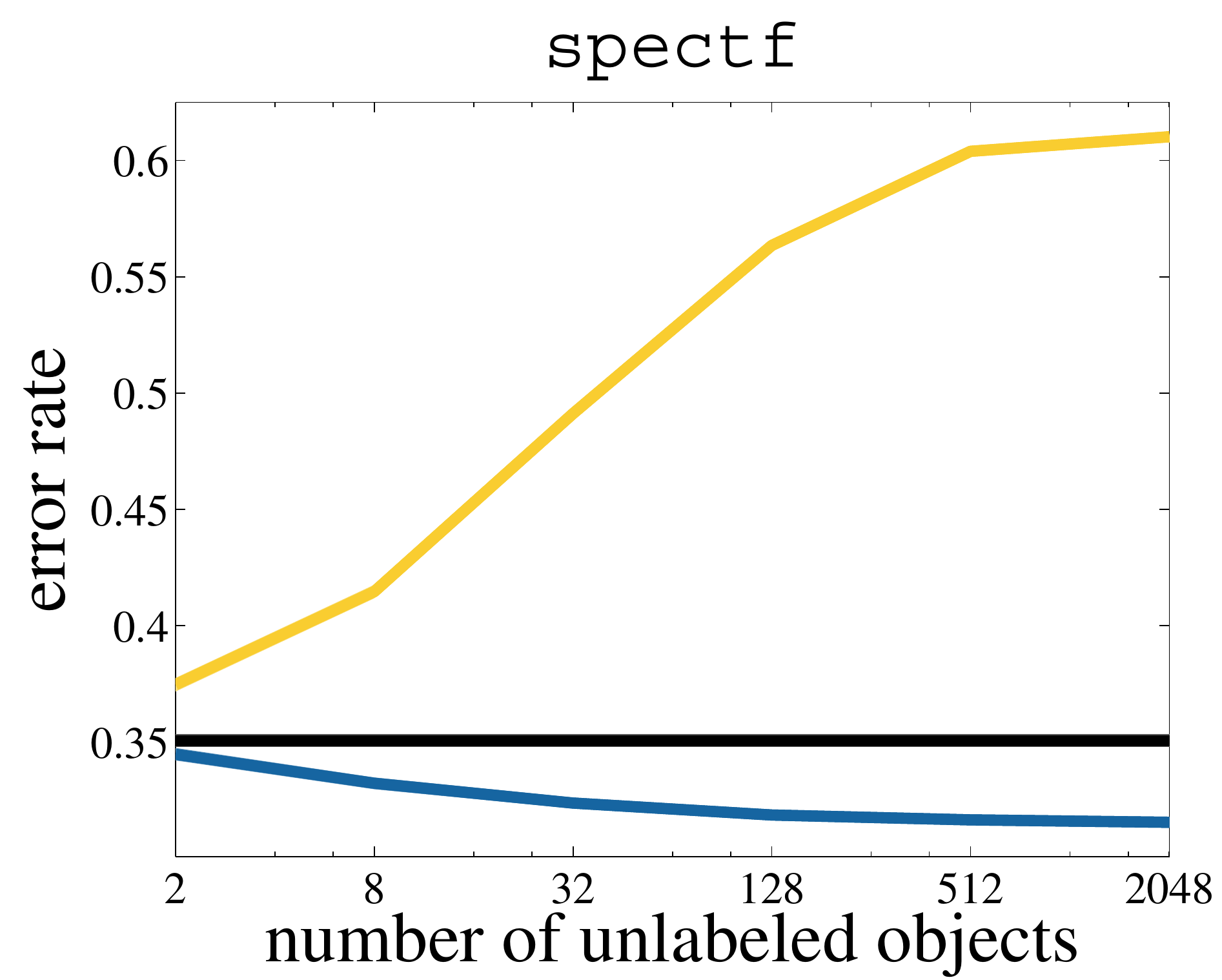}
\includegraphics[width=0.32\hsize]{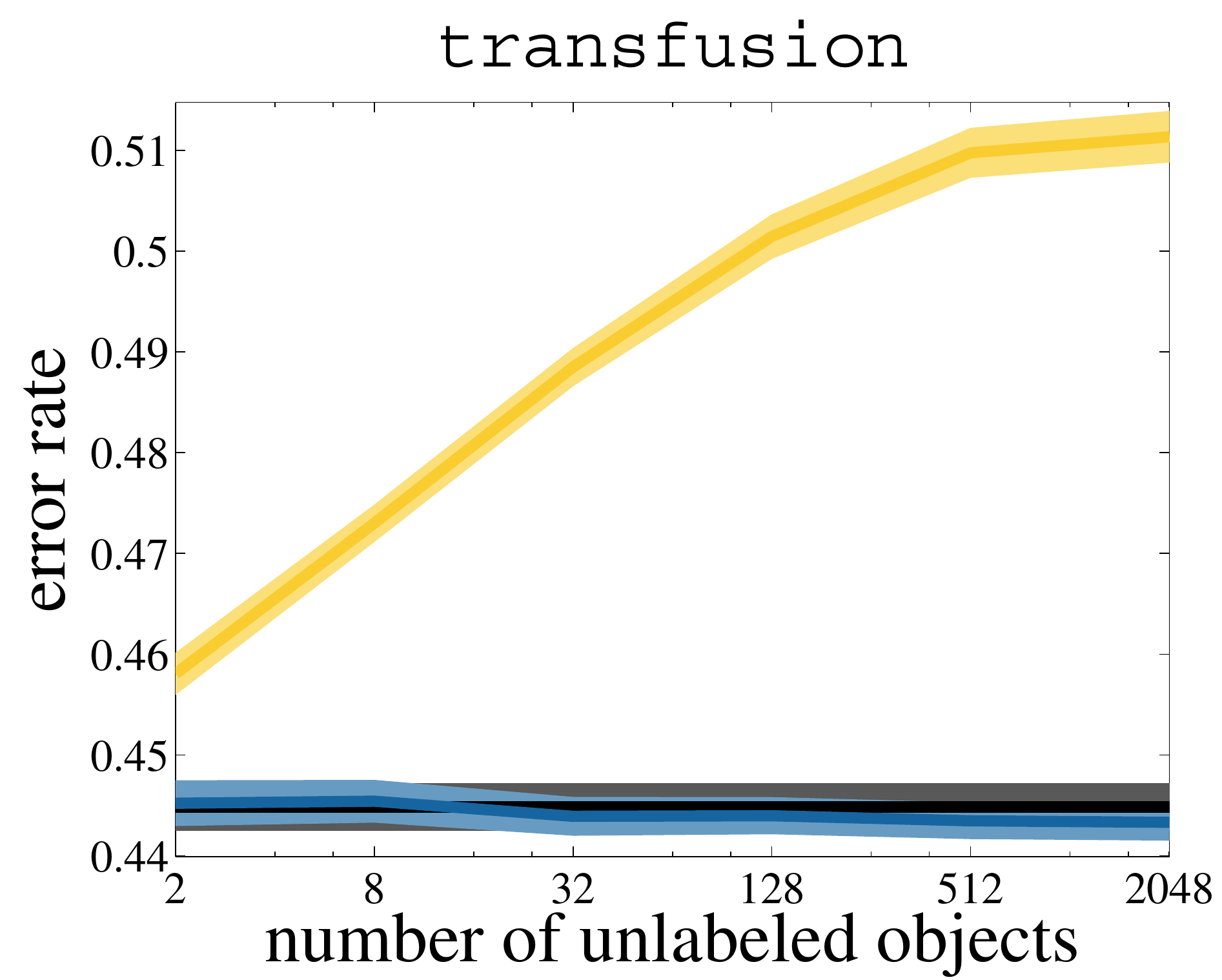}
\includegraphics[width=0.32\hsize]{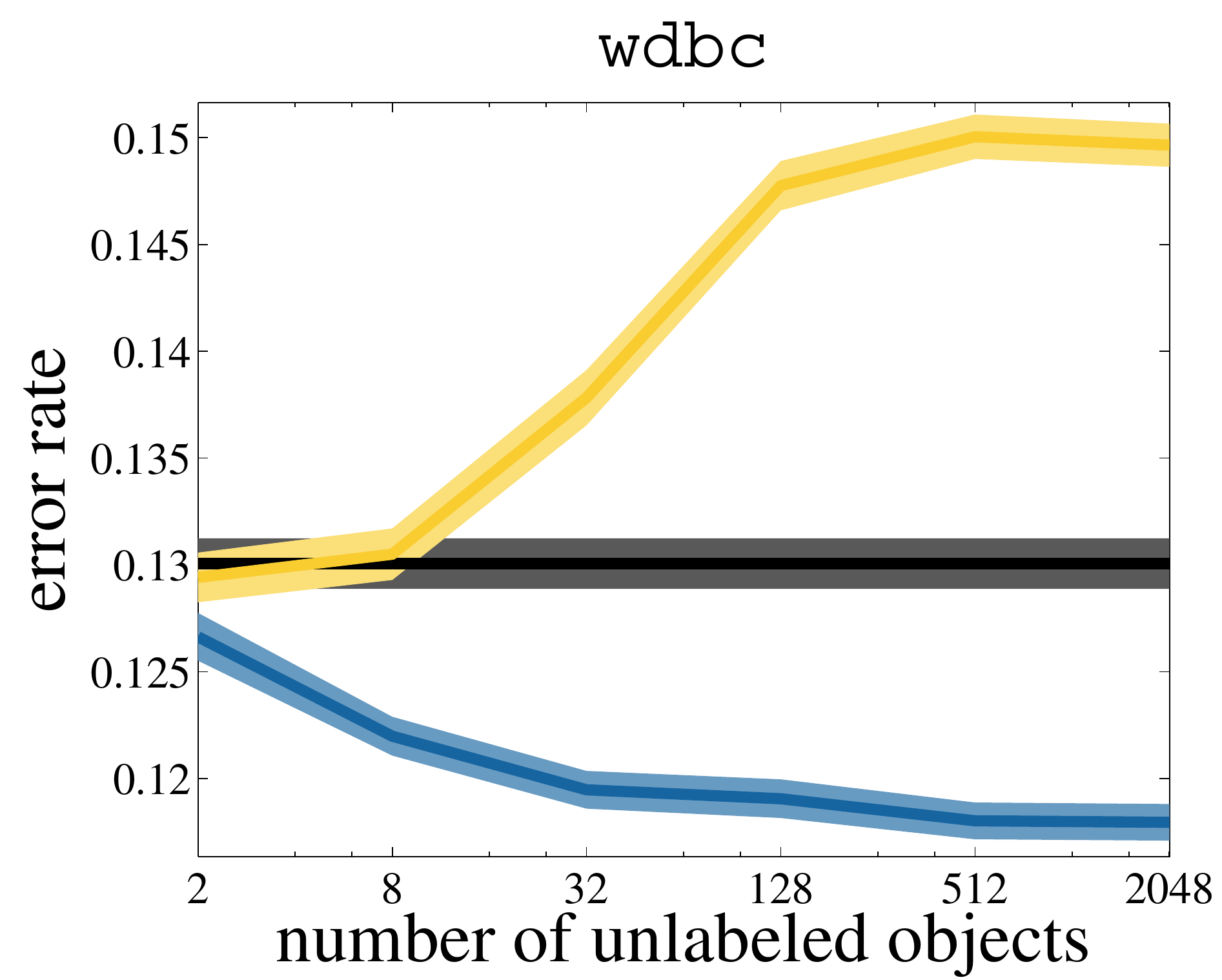}
\caption{Mean error rates for the supervised (black), self-learned (yellow), and the constrained NMC (blue) on the eight real-world data sets for various unlabeled sample sizes and a total of ten labeled training samples.}\label{fig:oneprime}
\end{figure*}


\begin{figure*}
\centering
\hrulefill~{\small NMC / log-likelihoods / 4 training samples}~\hrulefill \smallskip \\
\includegraphics[width=0.32\hsize]{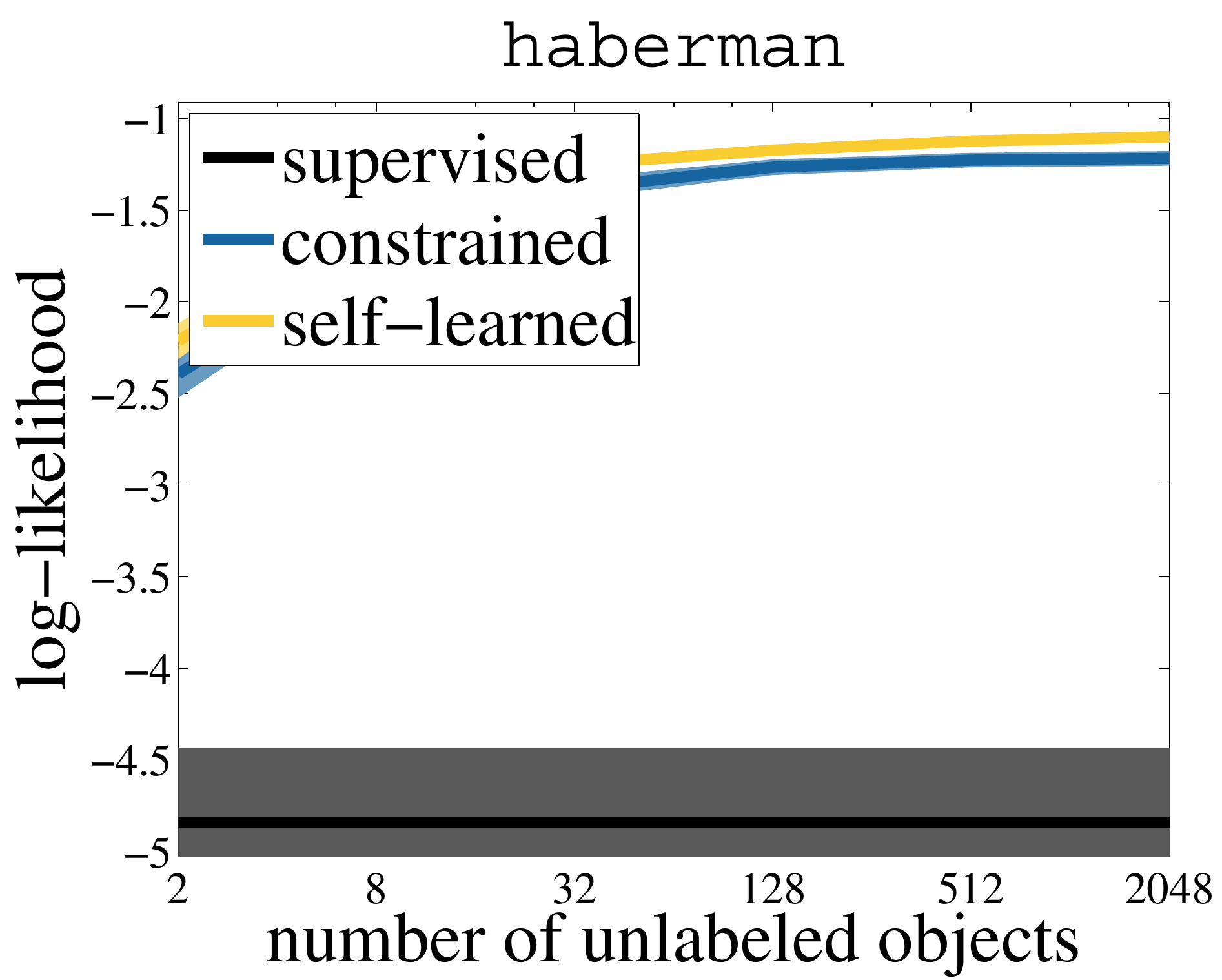}
\includegraphics[width=0.32\hsize]{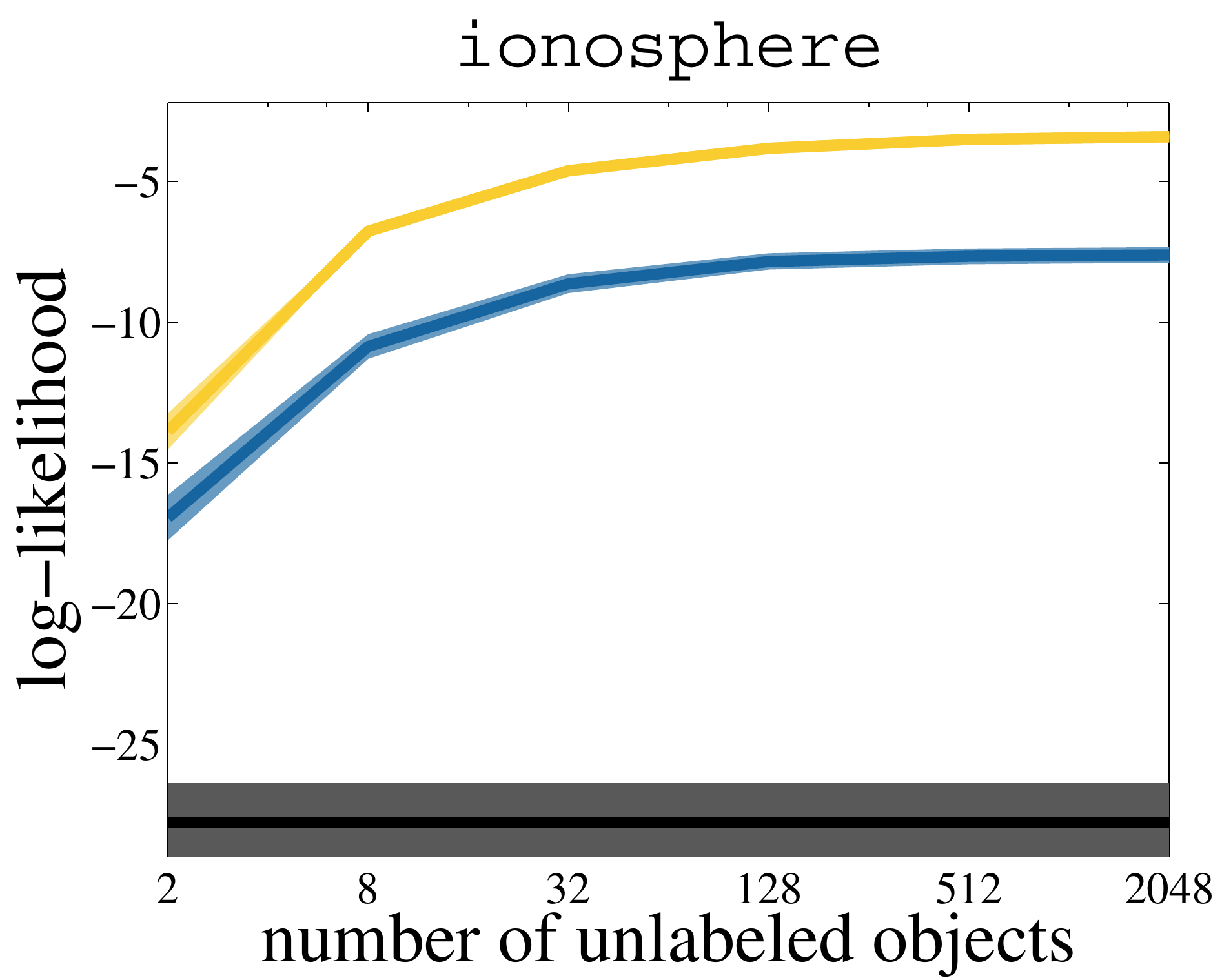}
\includegraphics[width=0.32\hsize]{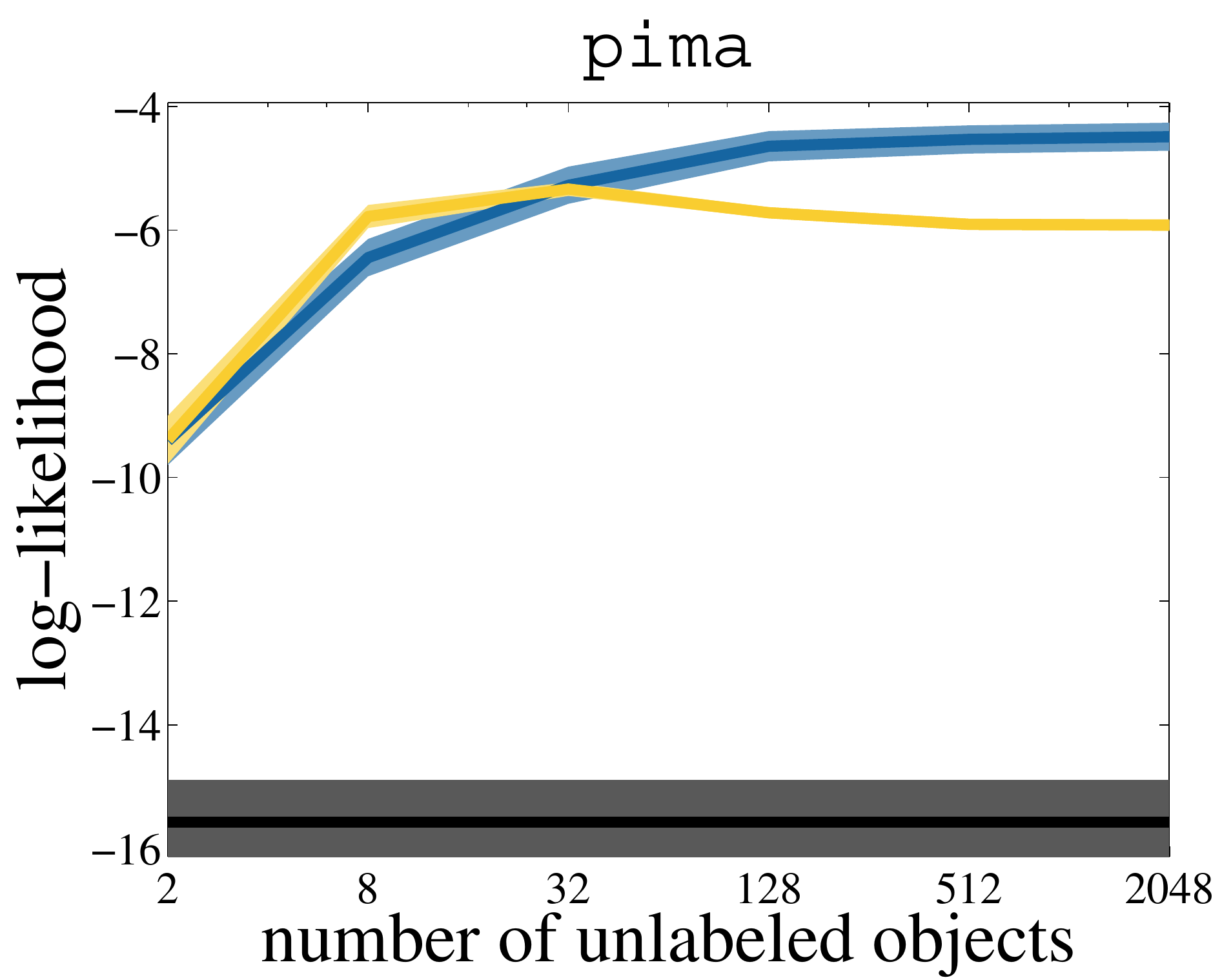} \bigskip \\
\includegraphics[width=0.32\hsize]{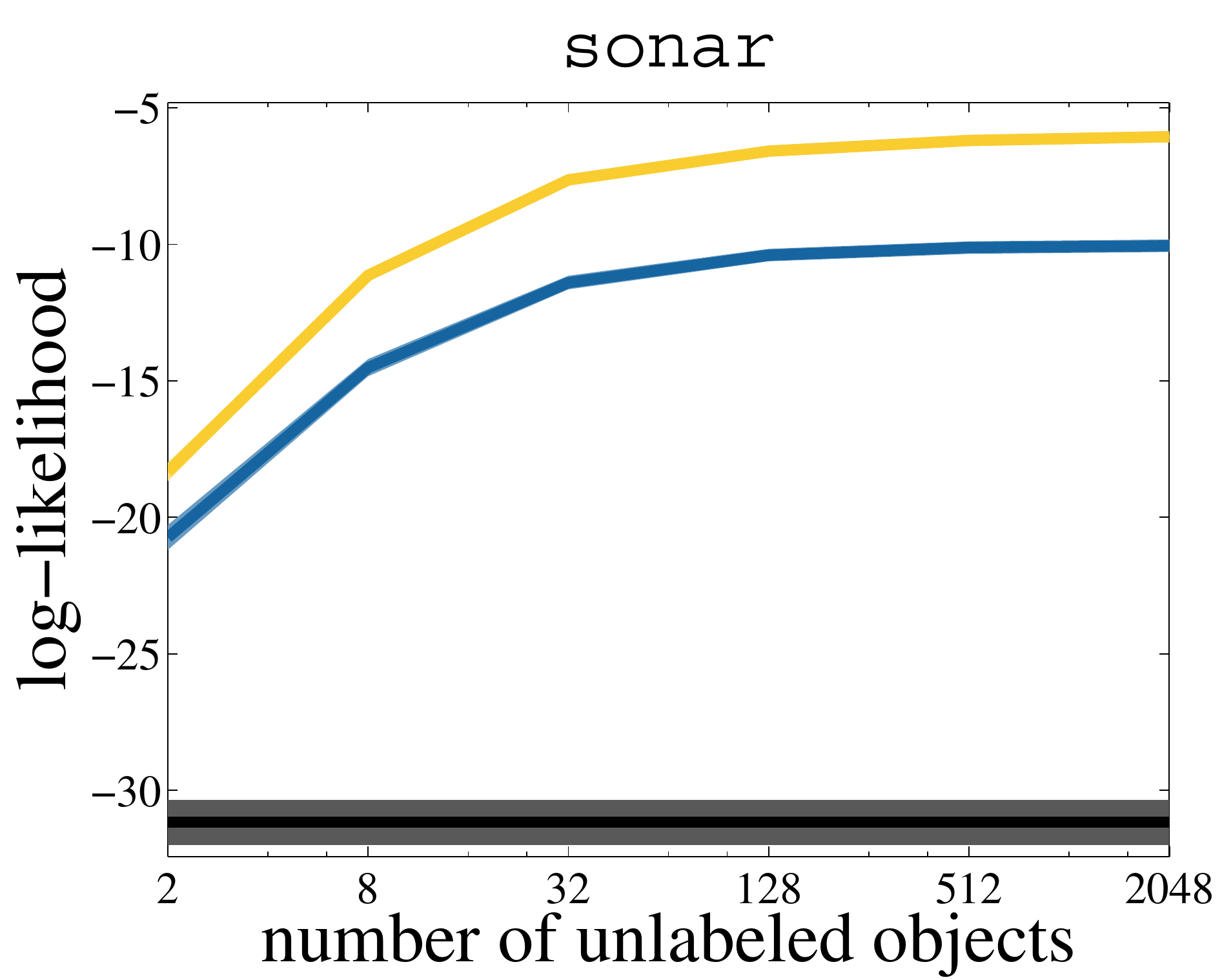}
\includegraphics[width=0.32\hsize]{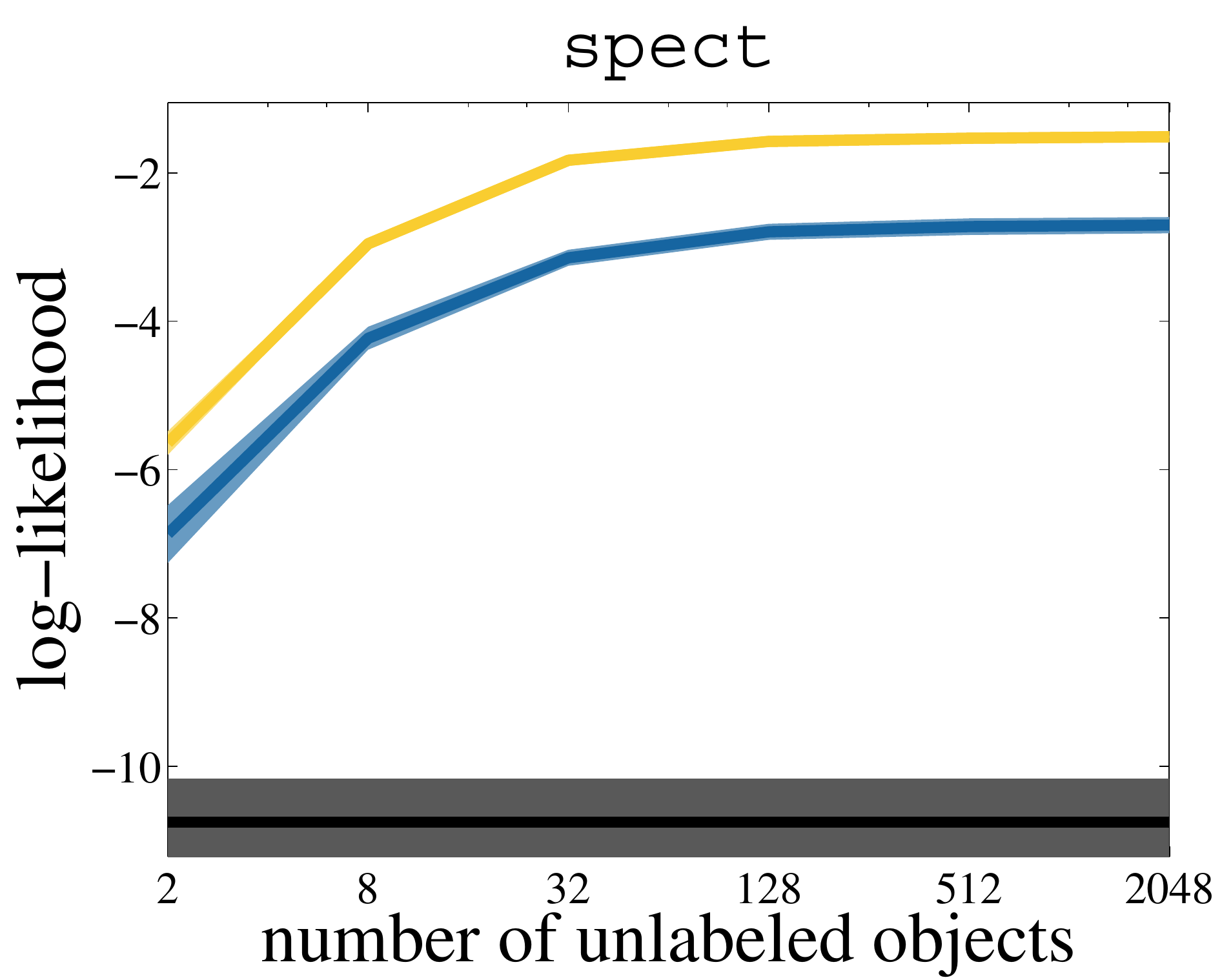} \bigskip \\
\includegraphics[width=0.32\hsize]{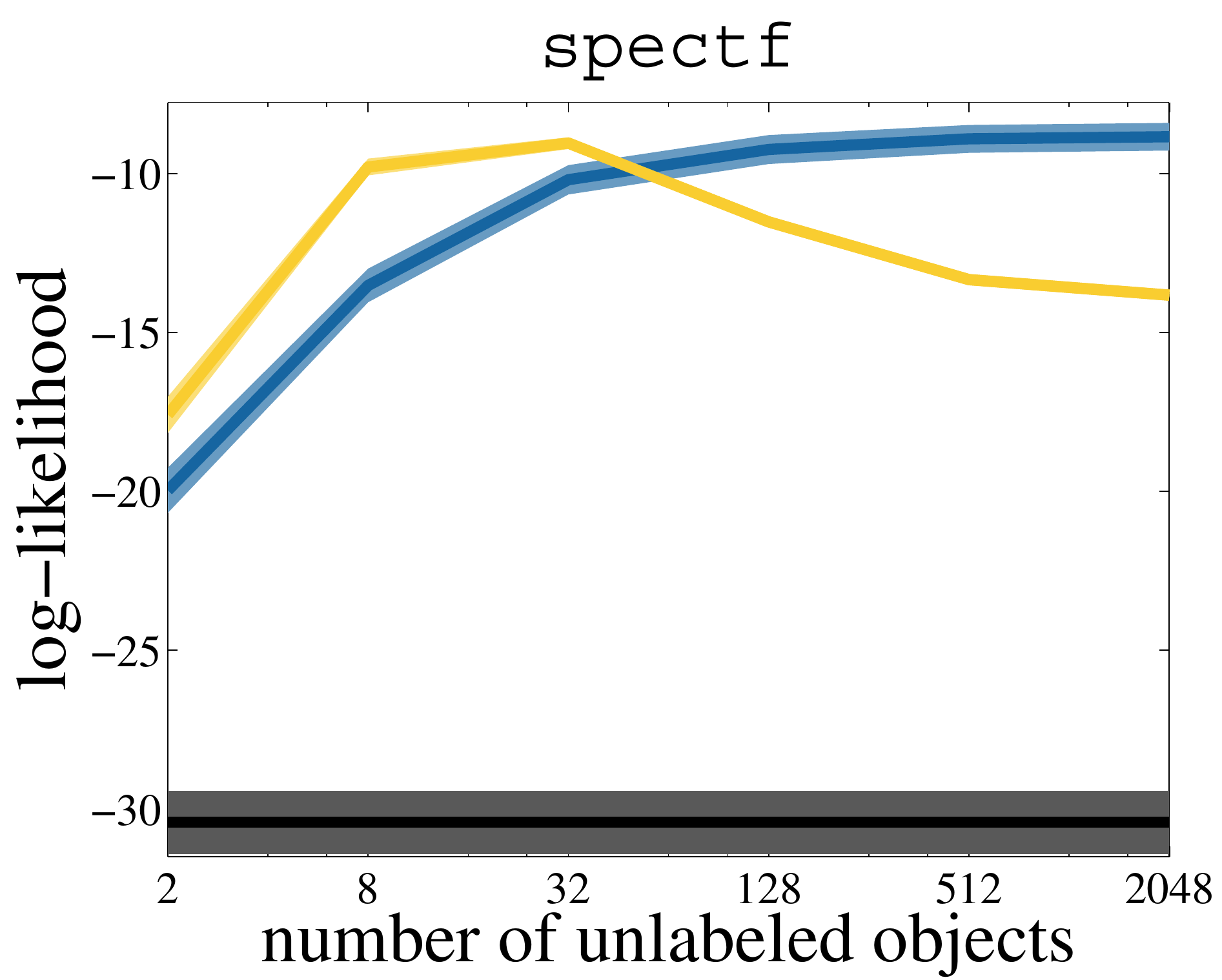}
\includegraphics[width=0.32\hsize]{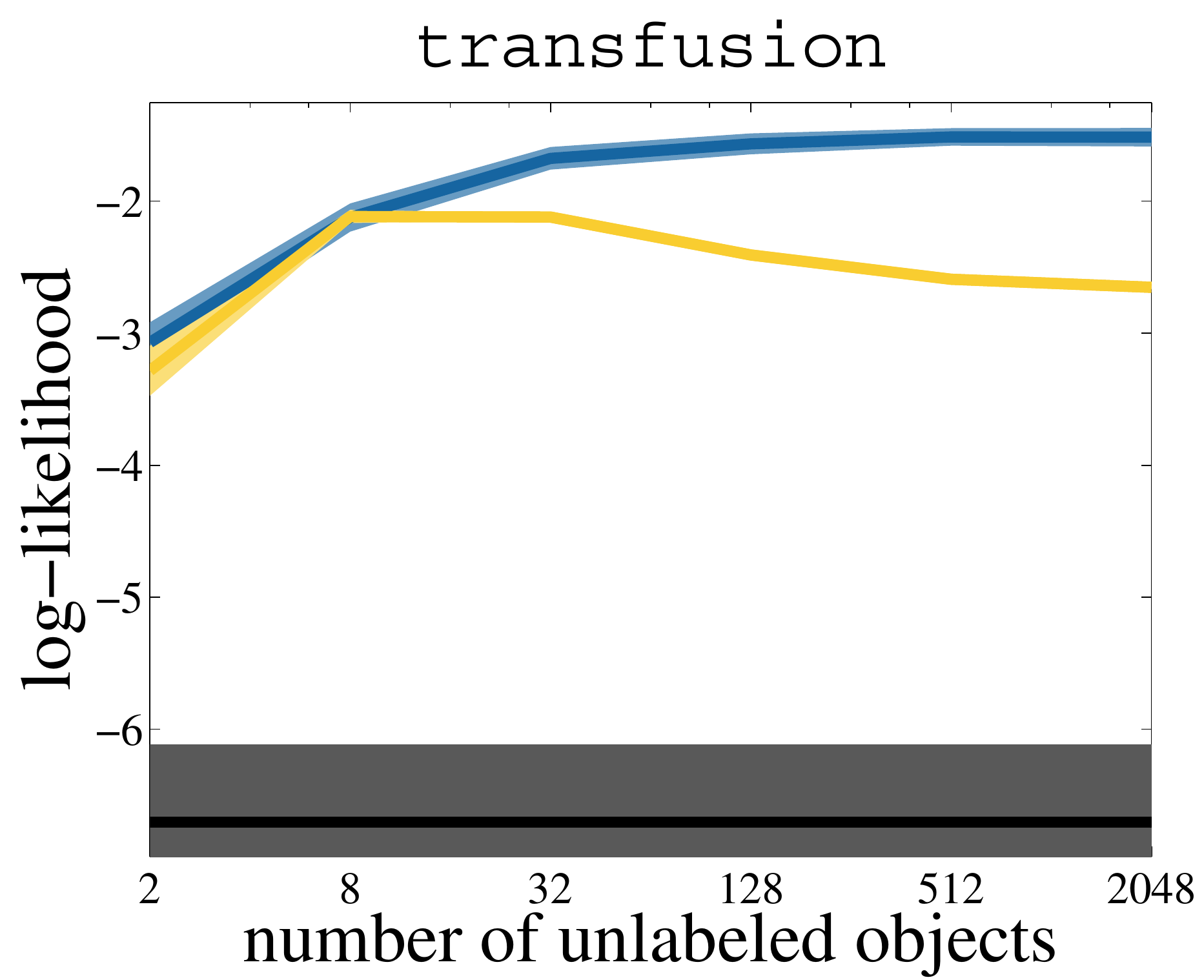}
\includegraphics[width=0.32\hsize]{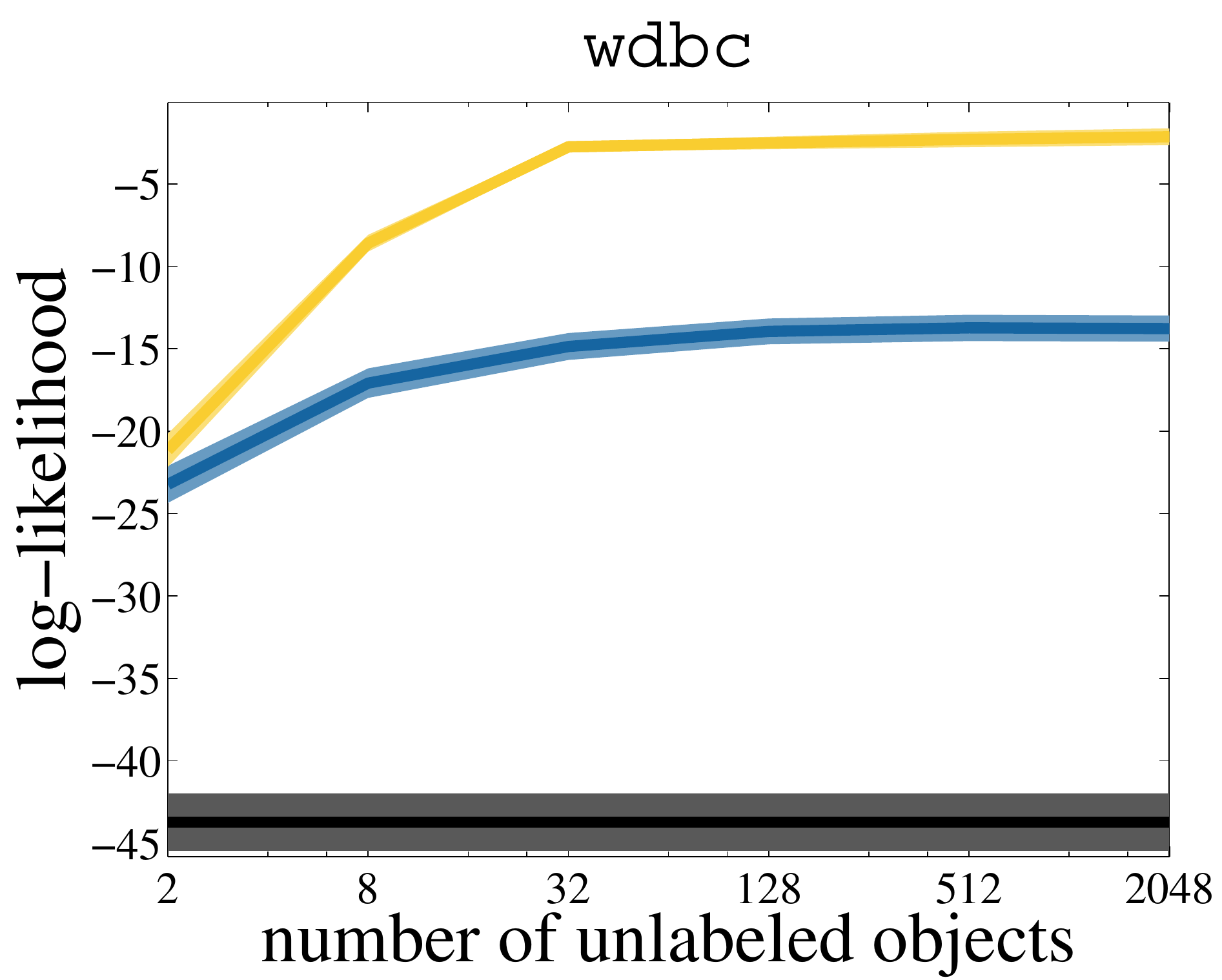}
\caption{Mean log-likelihood for the supervised (black), self-learned (yellow), and the constrained NMC (blue) on the eight real-world data sets for various unlabeled sample sizes and a total of four labeled training samples.  Compare these to the respective error rates in Figure \ref{fig:one}.}\label{fig:two}
\end{figure*}

\begin{figure*}
\centering
\hrulefill~{\small NMC / log-likelihoods / 10 training samples~}\hrulefill \smallskip \\
\includegraphics[width=0.32\hsize]{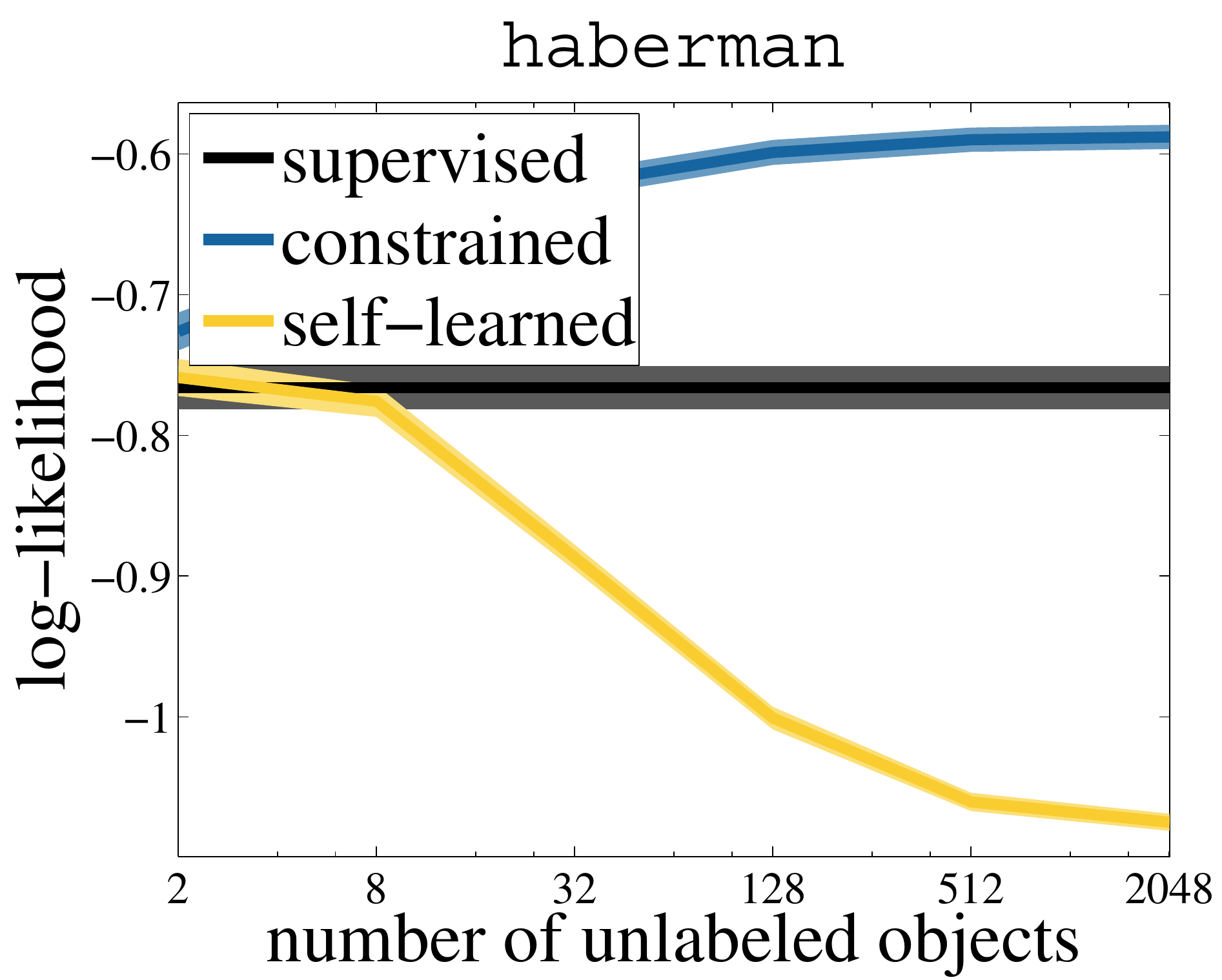}
\includegraphics[width=0.32\hsize]{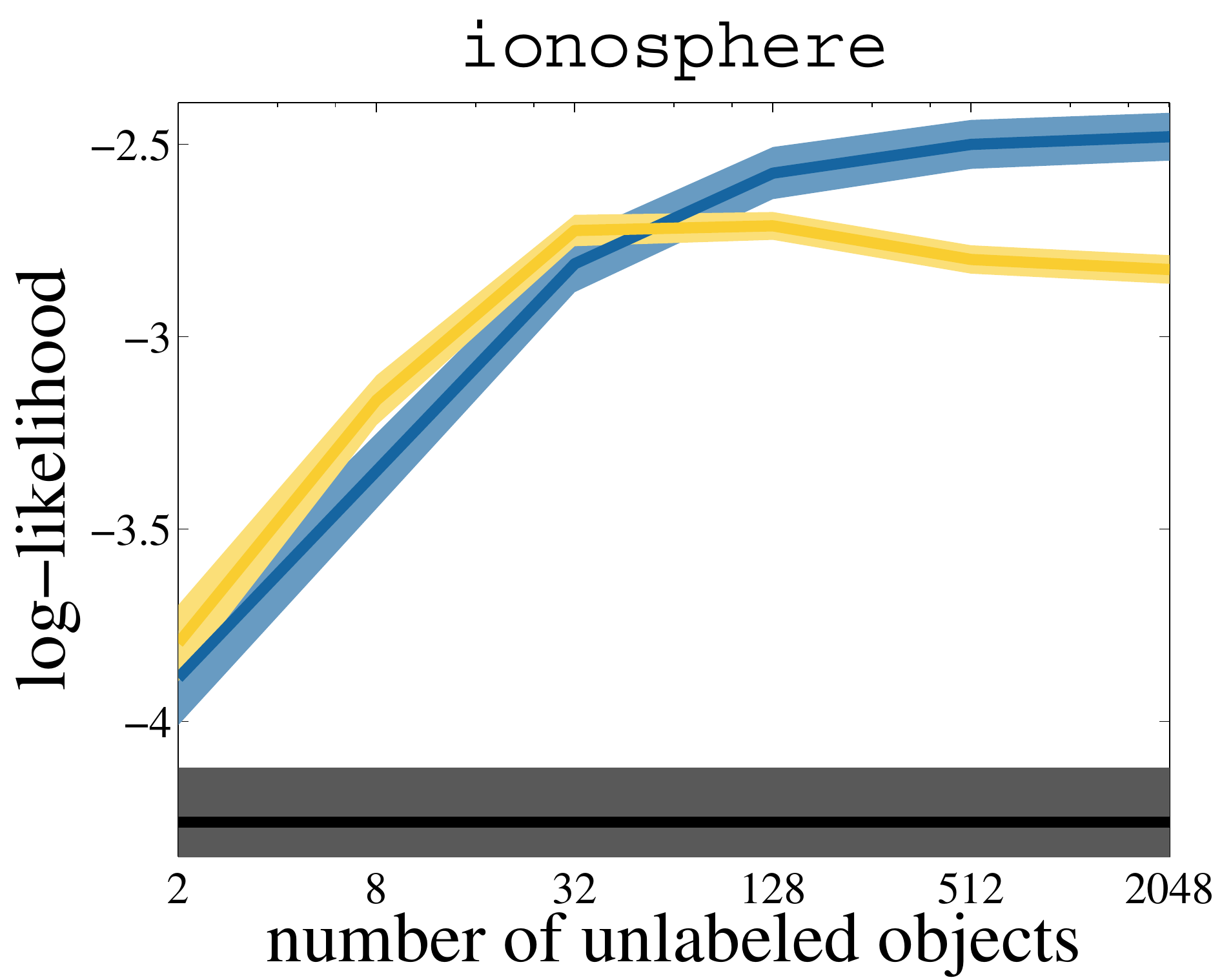}
\includegraphics[width=0.32\hsize]{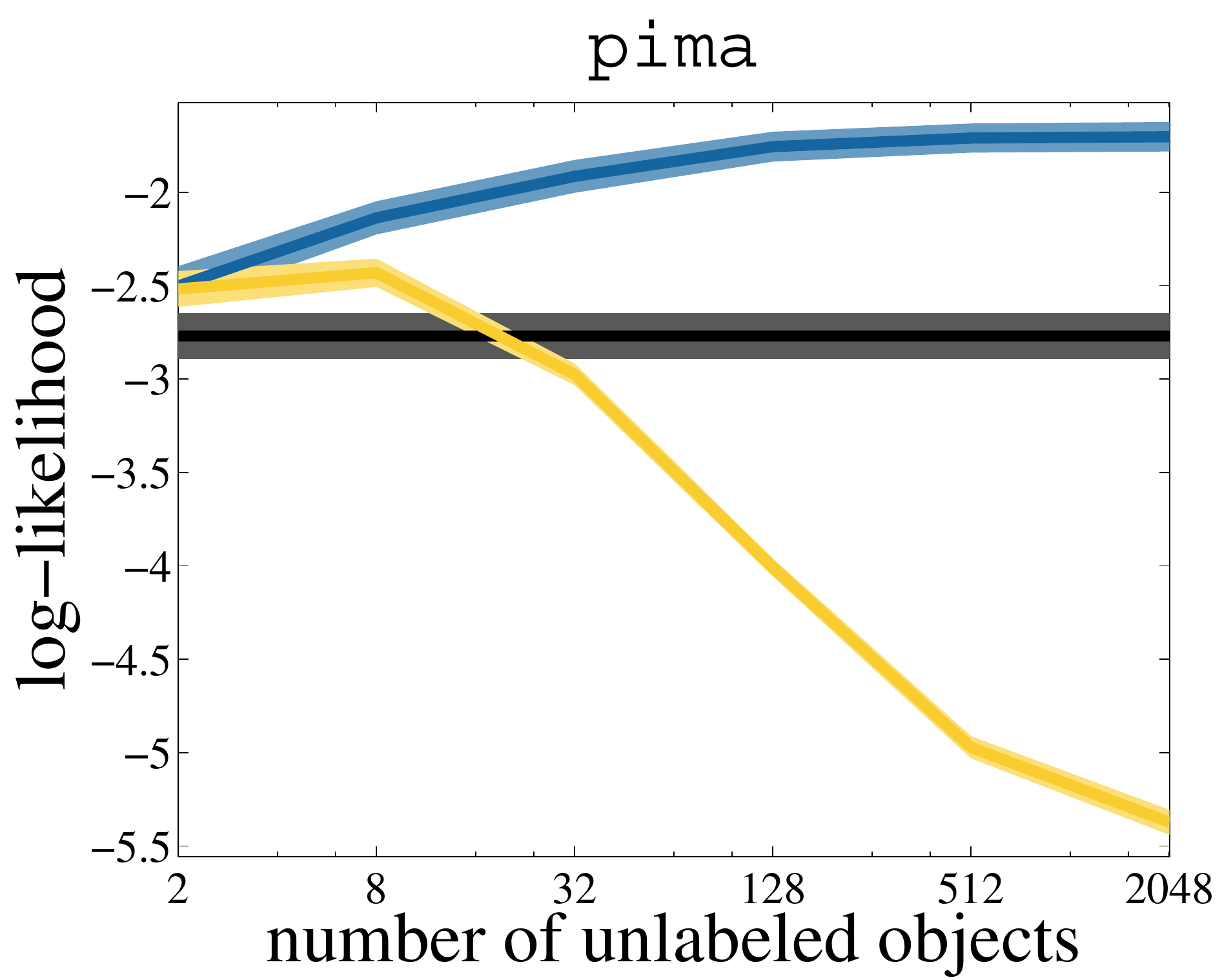} \bigskip \\
\includegraphics[width=0.32\hsize]{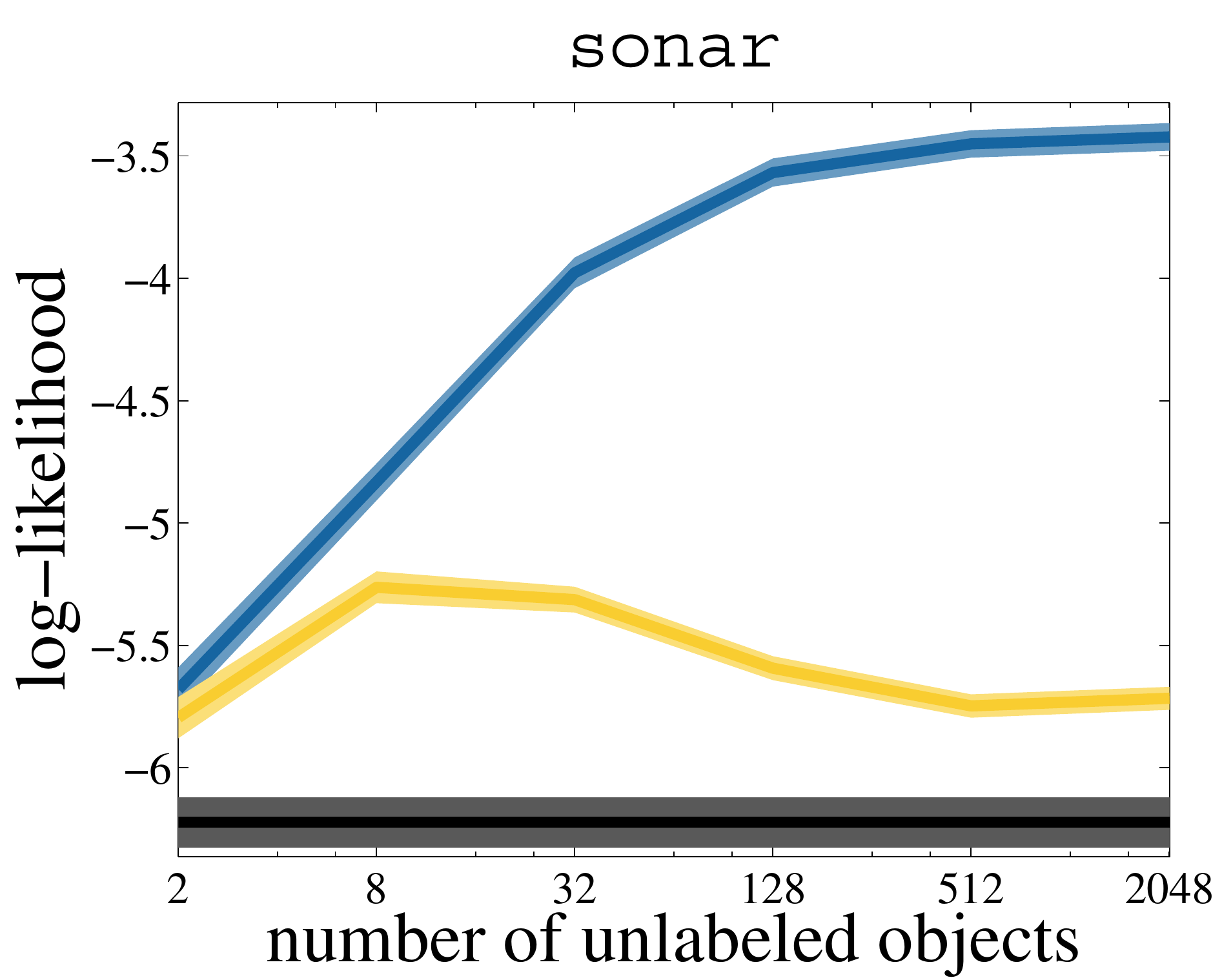}
\includegraphics[width=0.32\hsize]{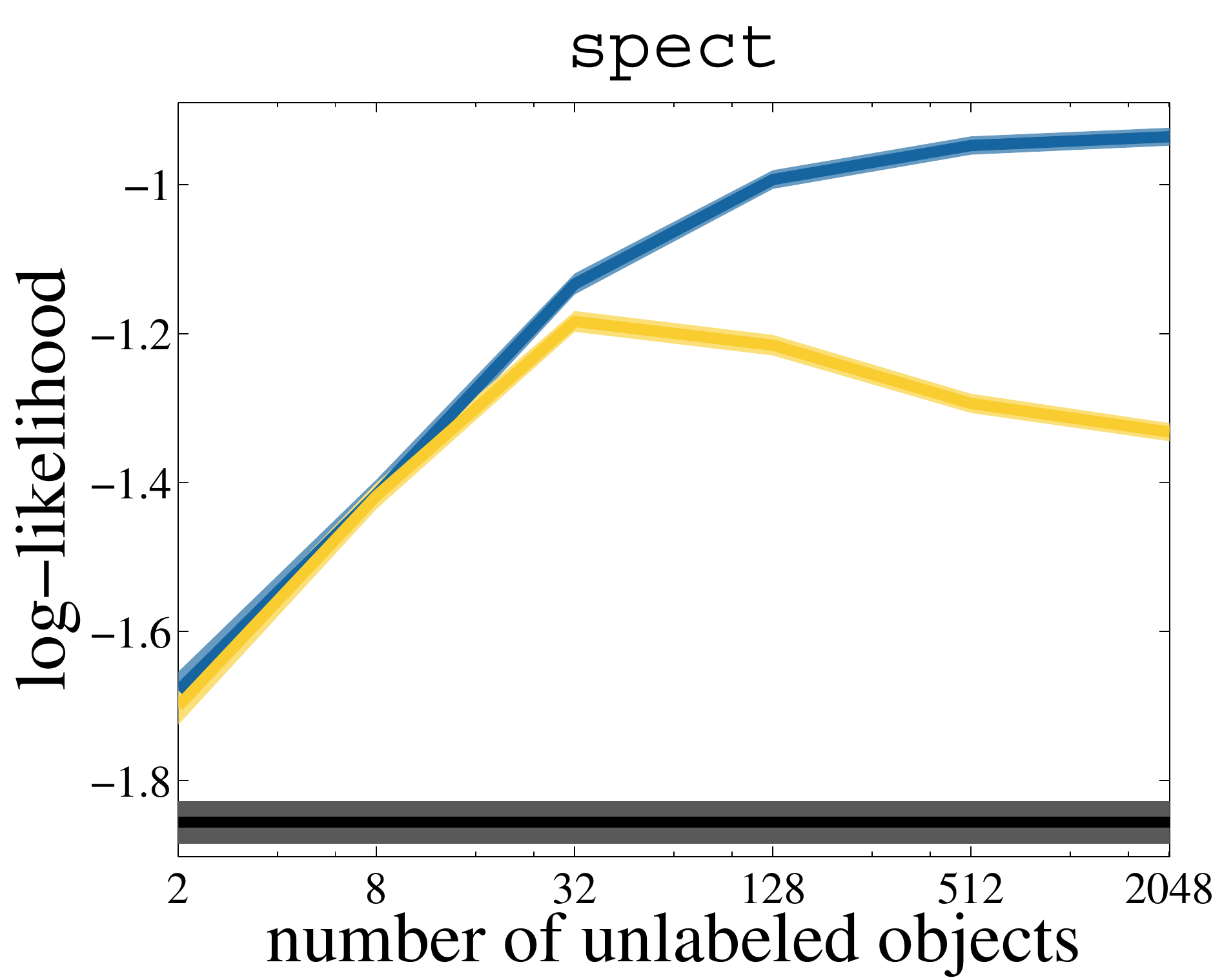} \bigskip \\
\includegraphics[width=0.32\hsize]{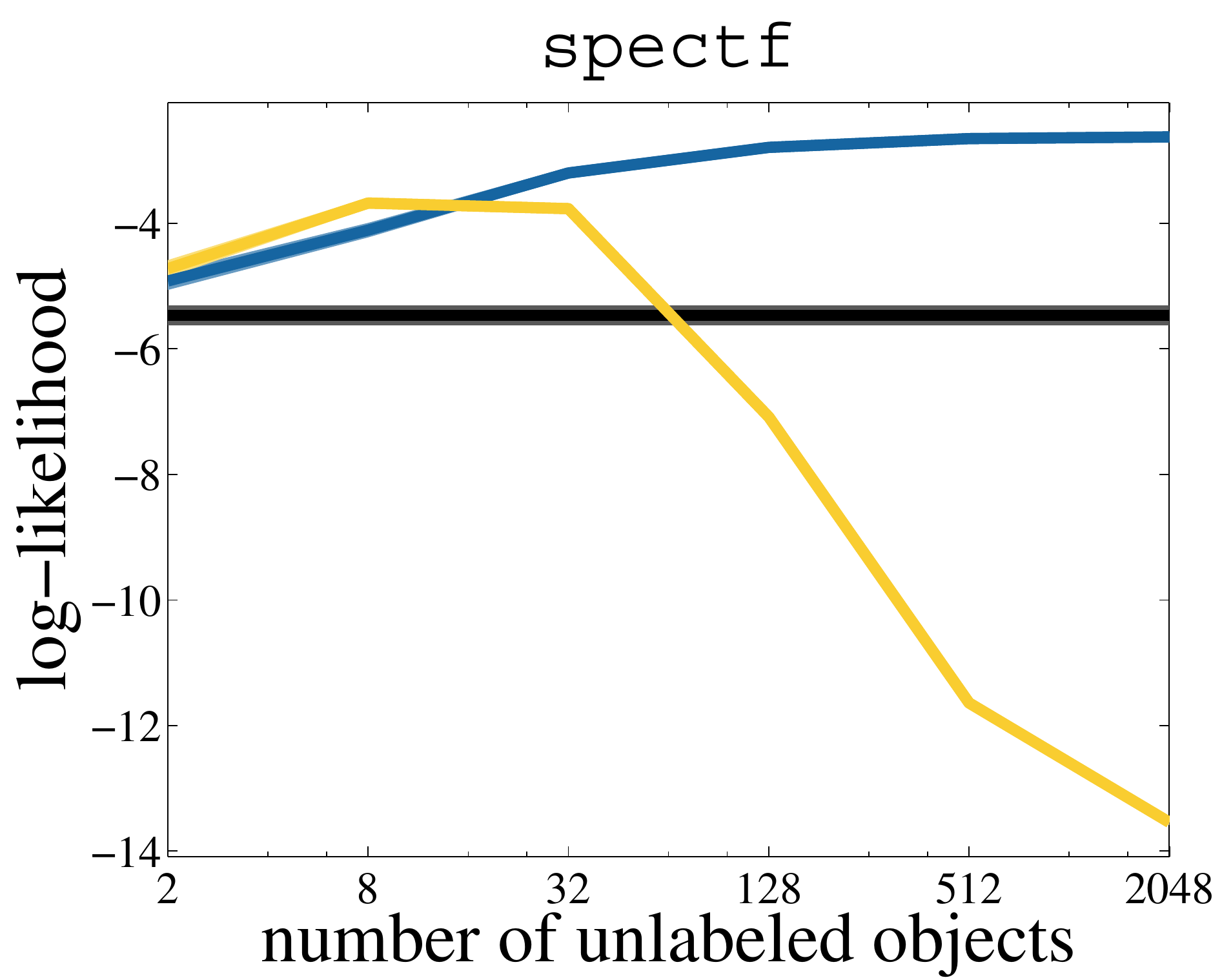}
\includegraphics[width=0.32\hsize]{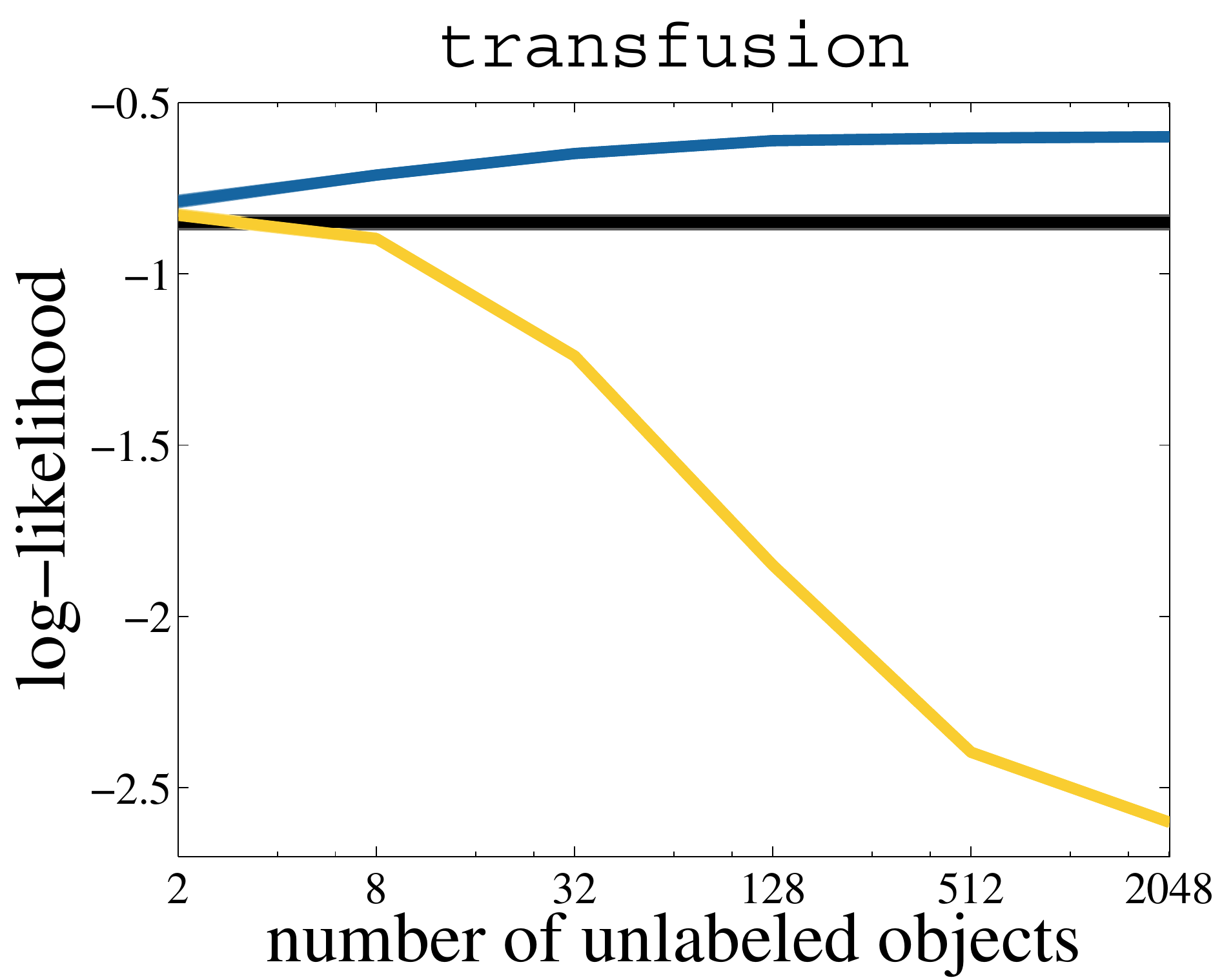}
\includegraphics[width=0.32\hsize]{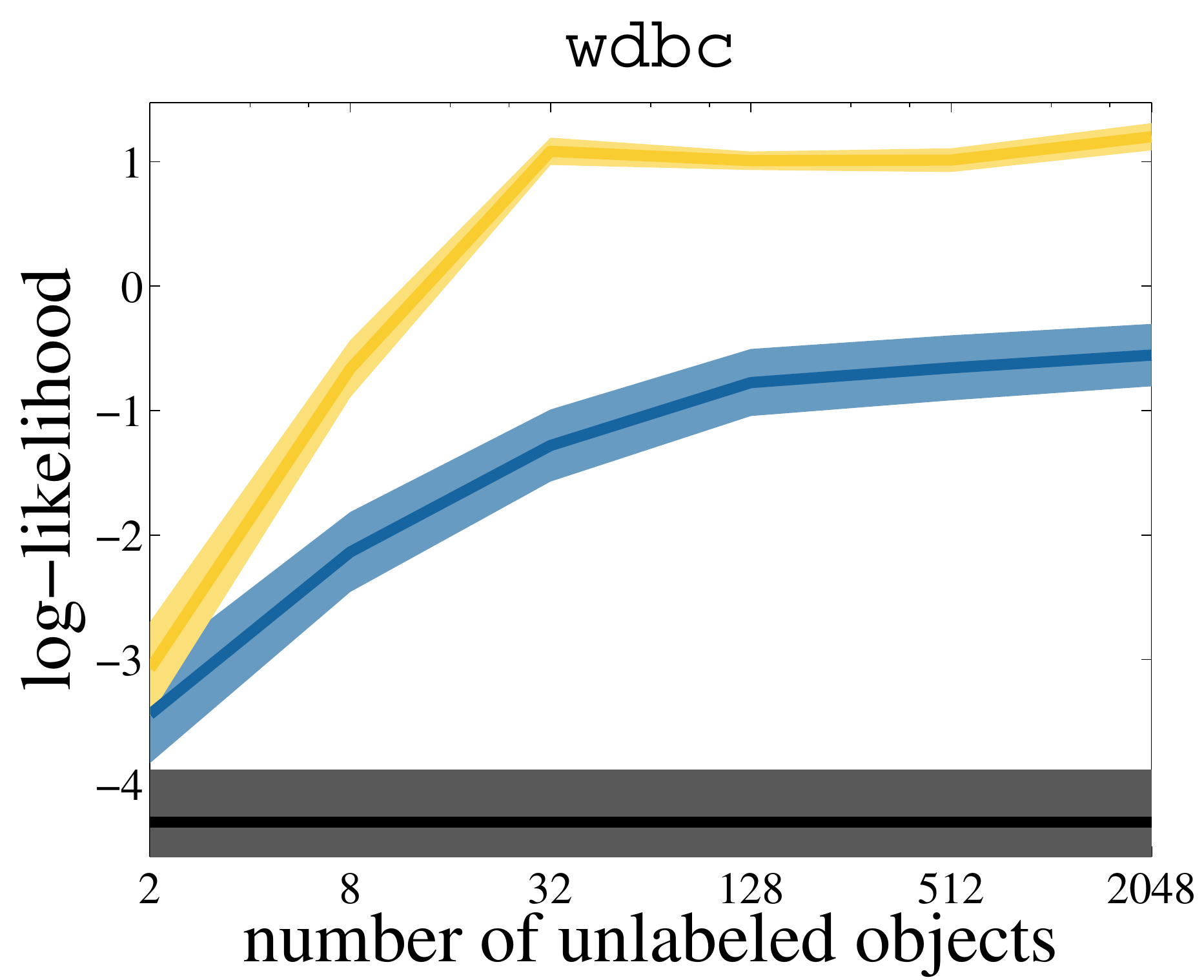}
\caption{Mean log-likelihood for the supervised (black), self-learned (yellow), and the constrained NMC (blue) on the eight real-world data sets for various unlabeled sample sizes and a total of ten labeled training samples.   Compare these to the respective error rates in Figure \ref{fig:oneprime}.}\label{fig:twoprime}
\end{figure*}


\begin{figure*}
\centering
\hrulefill~{\small LDA / error rates / 100 training samples}~\hrulefill \smallskip \\
\includegraphics[width=0.32\hsize]{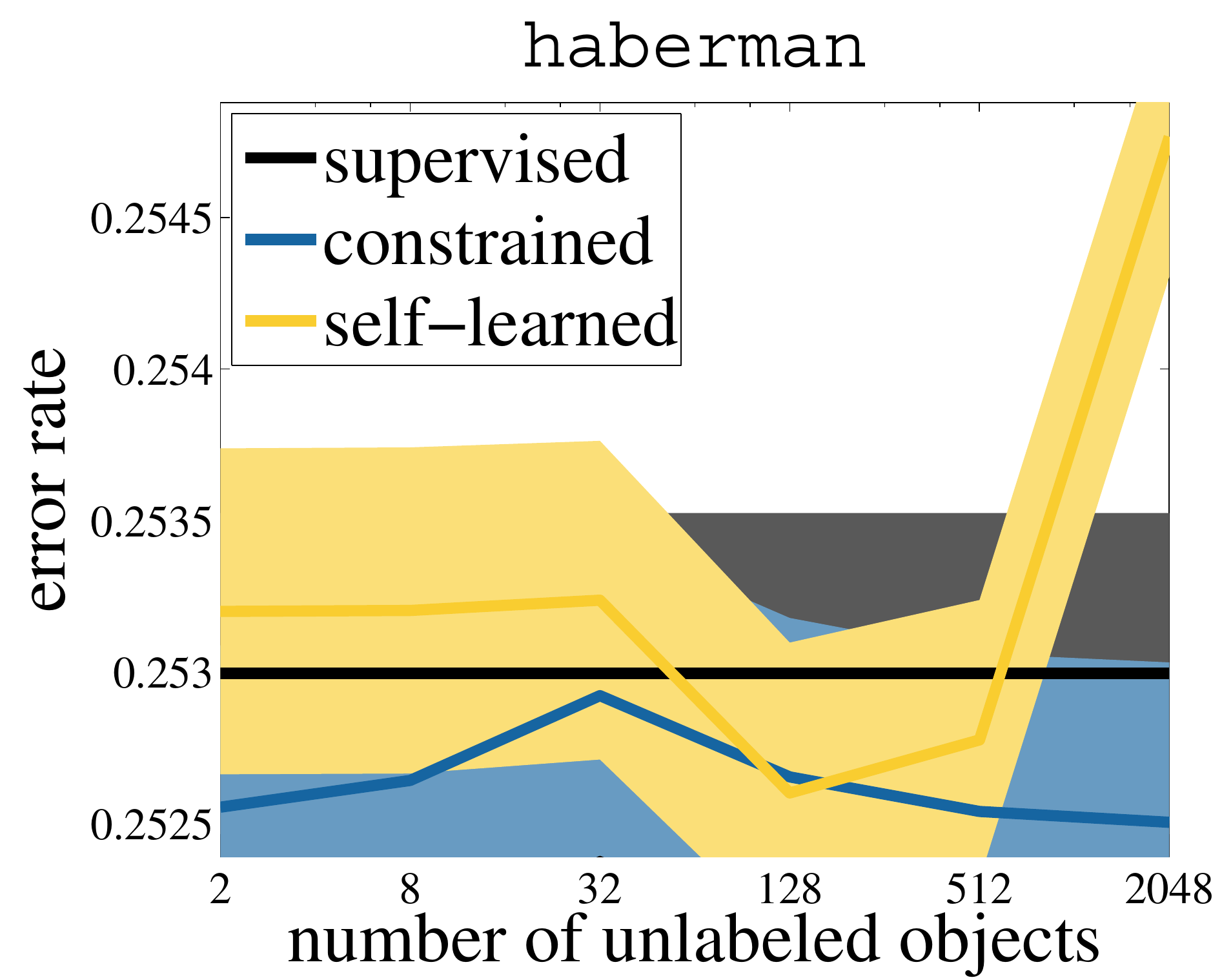}
\includegraphics[width=0.32\hsize]{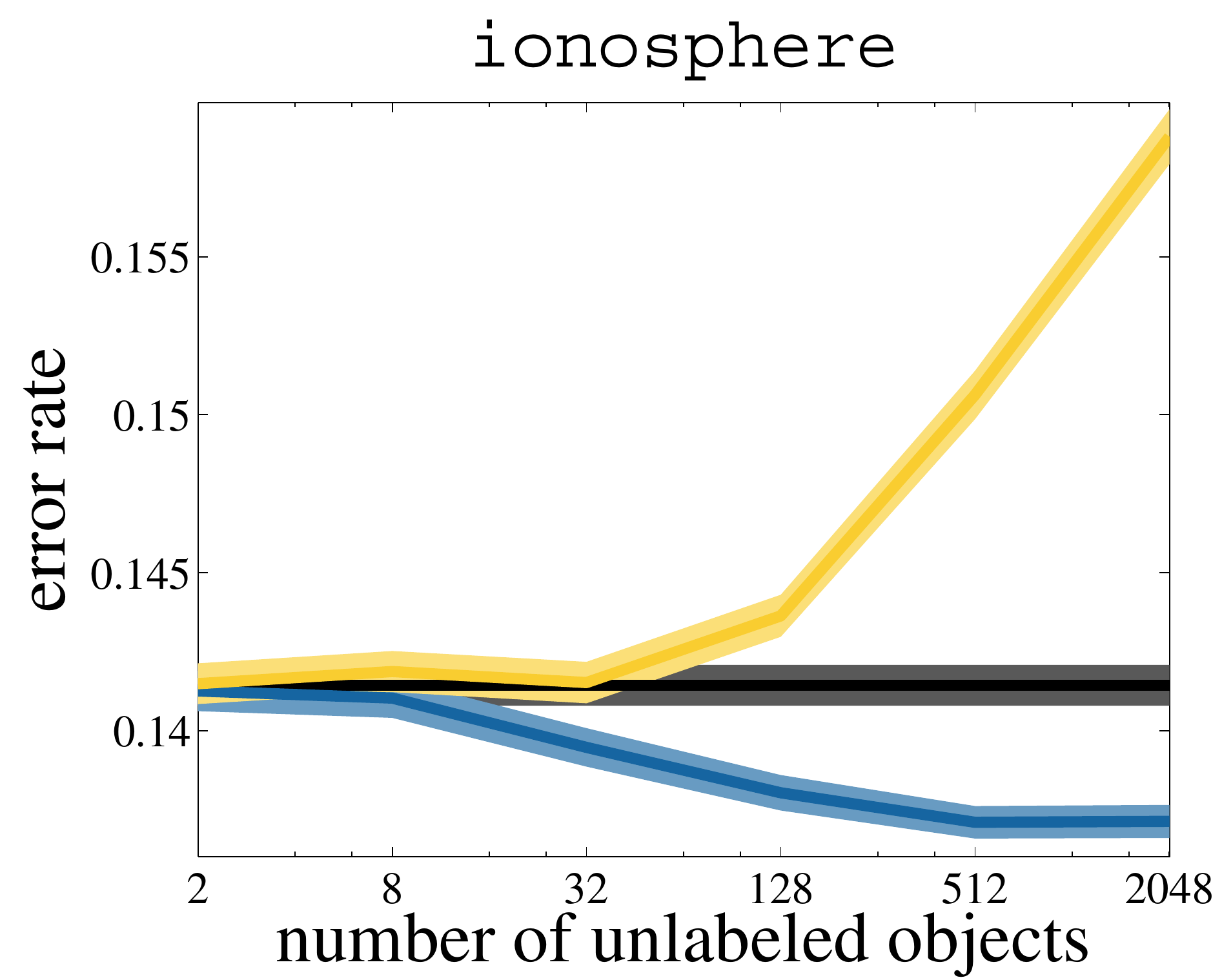}
\includegraphics[width=0.32\hsize]{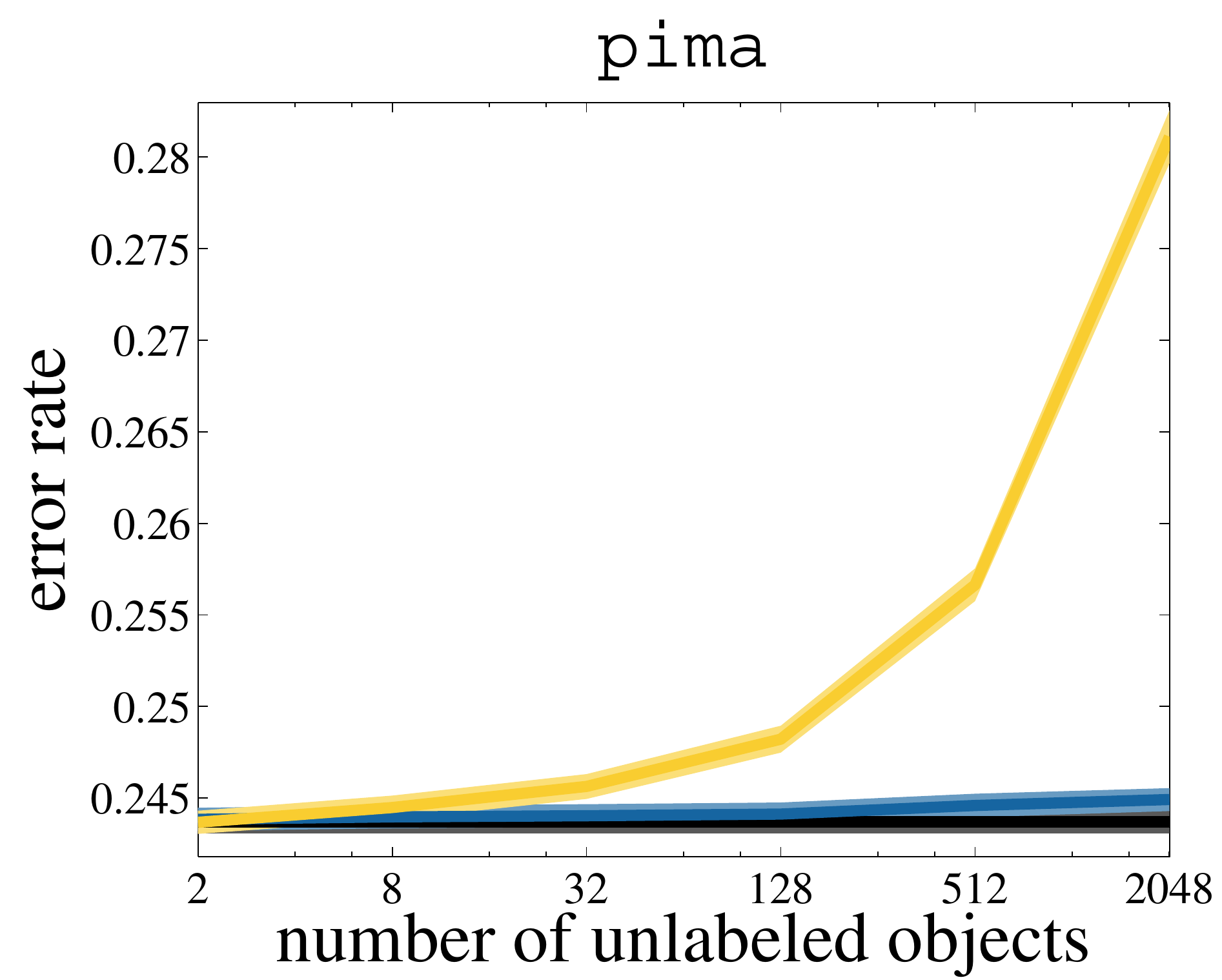} \bigskip \\
\includegraphics[width=0.32\hsize]{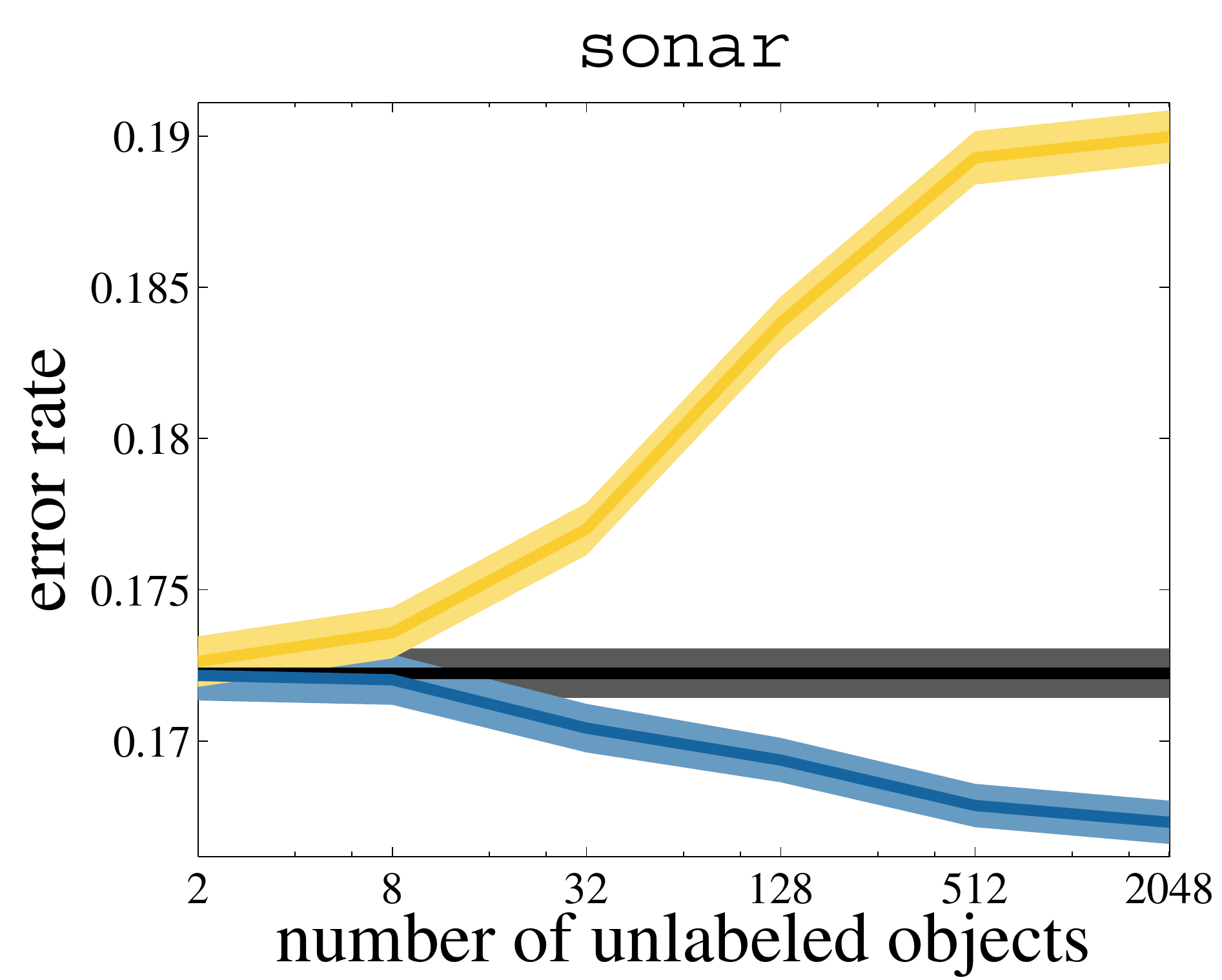}
\includegraphics[width=0.32\hsize]{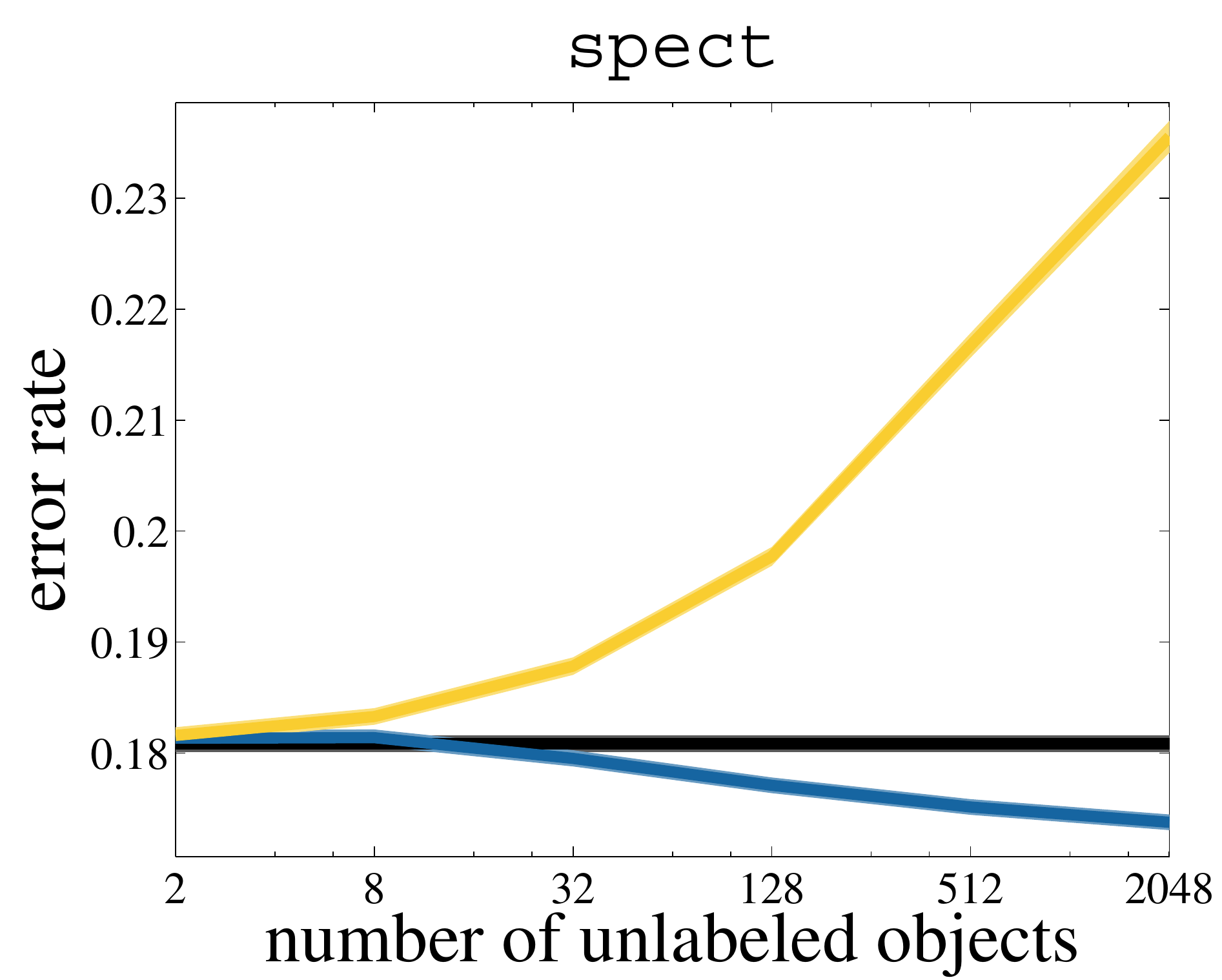} \bigskip \\
\includegraphics[width=0.32\hsize]{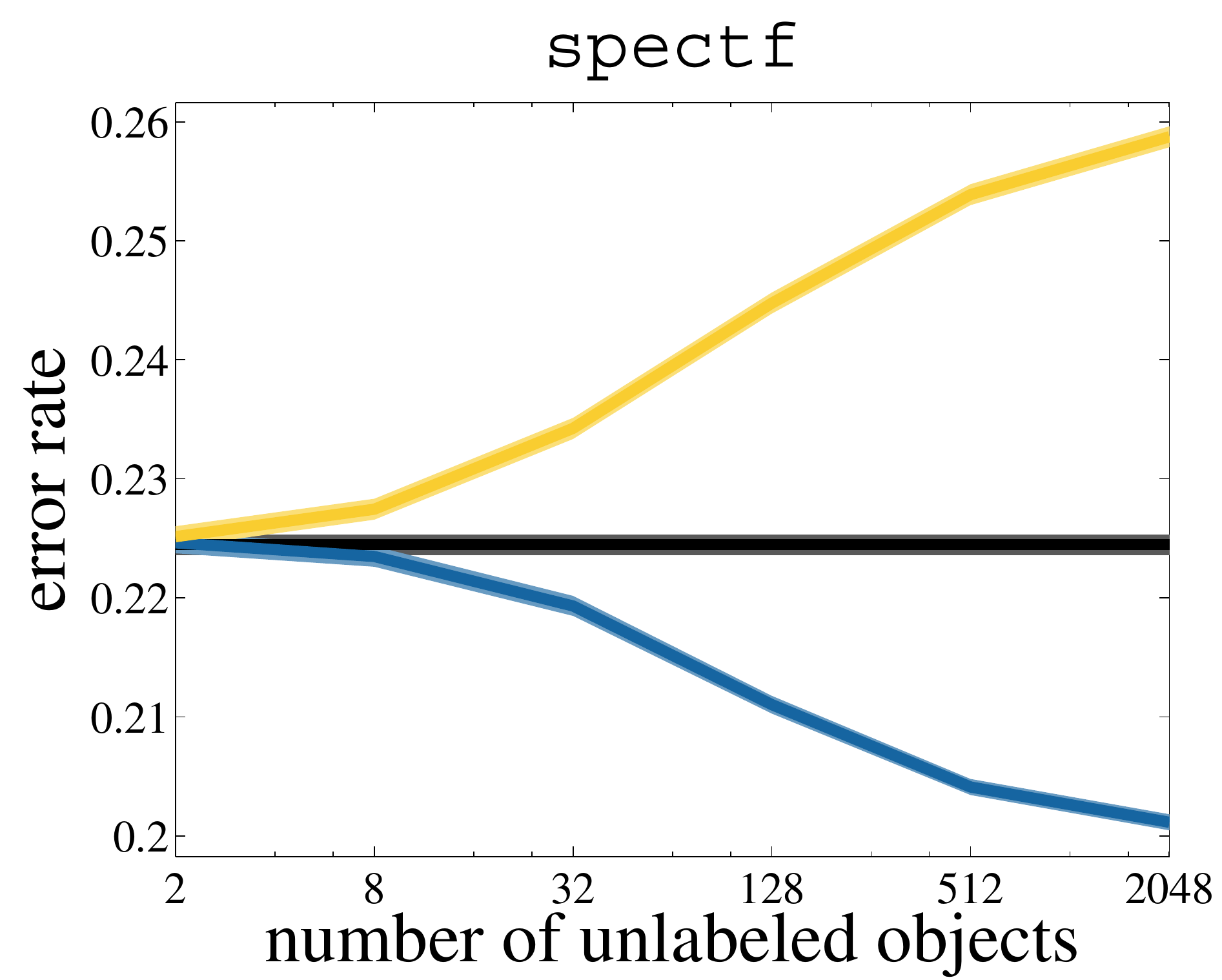}
\includegraphics[width=0.32\hsize]{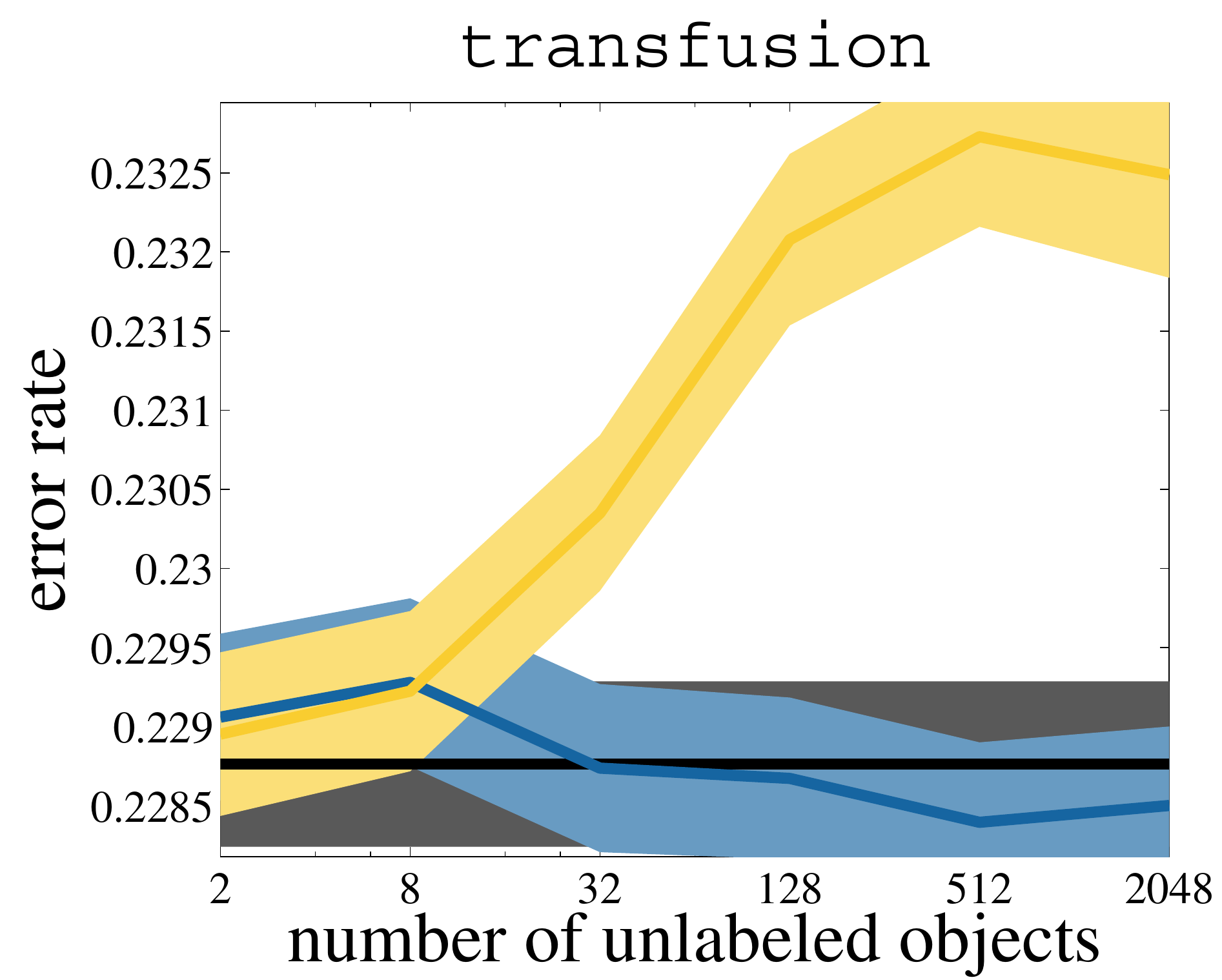}
\includegraphics[width=0.32\hsize]{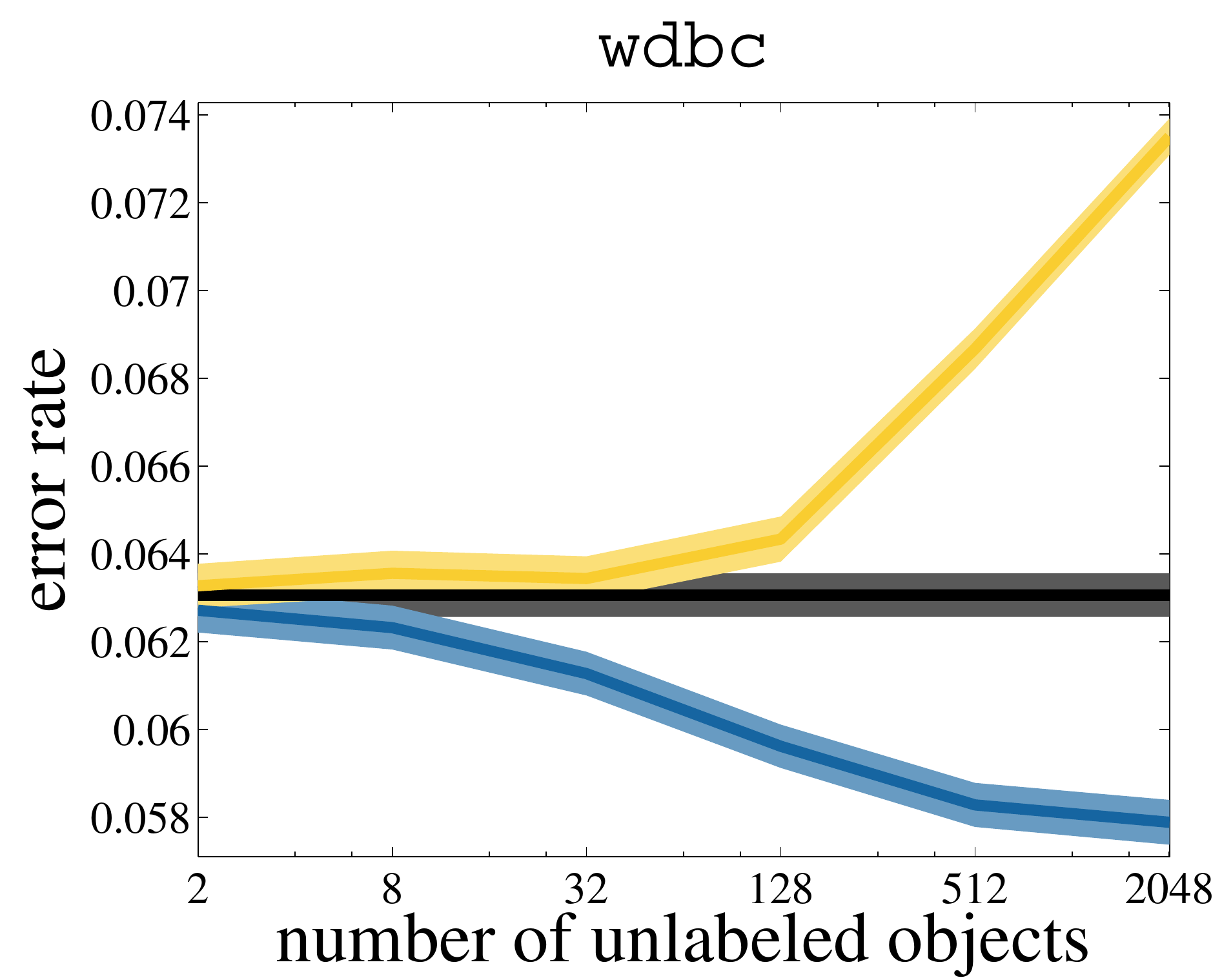}
\caption{Mean error rates for supervised (black), self-learned (yellow), and constrained LDA (blue) on the eight real-world data sets for various unlabeled sample sizes and a total of 100 labeled training samples.}\label{fig:five}
\end{figure*}

\begin{figure*}
\centering
\hrulefill~{\small LDA / log-likelihoods / 100 training samples}~\hrulefill \smallskip \\
\includegraphics[width=0.32\hsize]{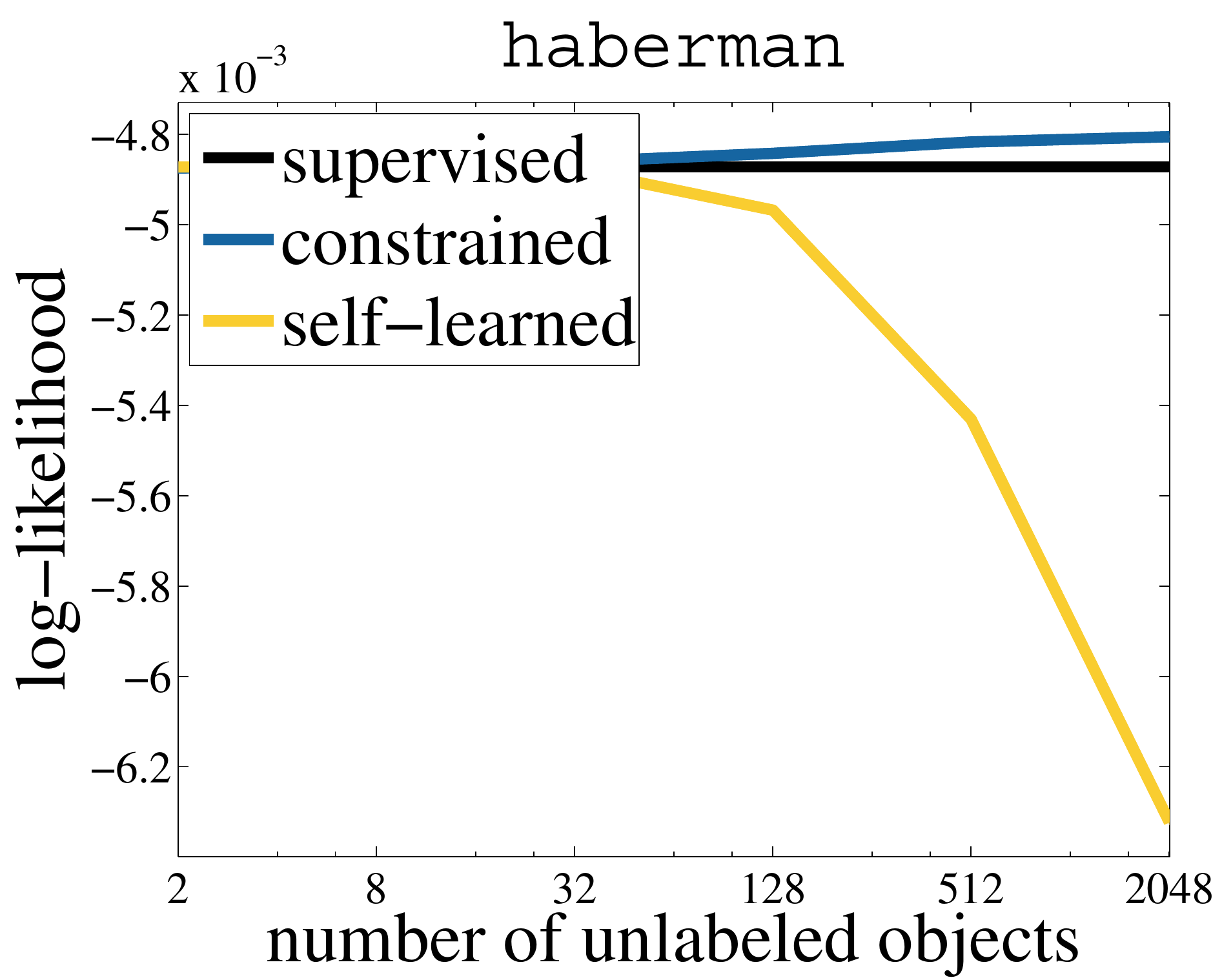}
\includegraphics[width=0.32\hsize]{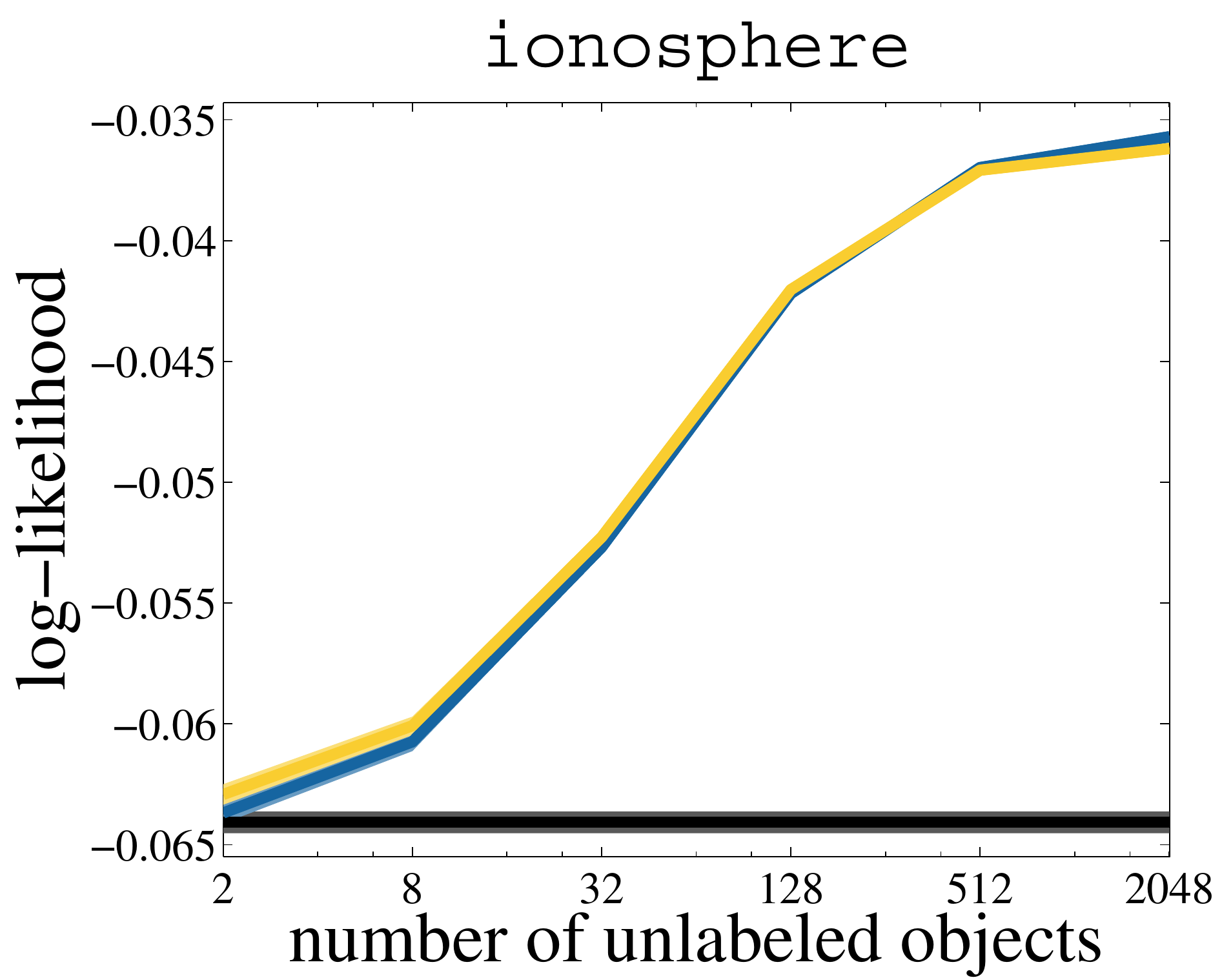}
\includegraphics[width=0.32\hsize]{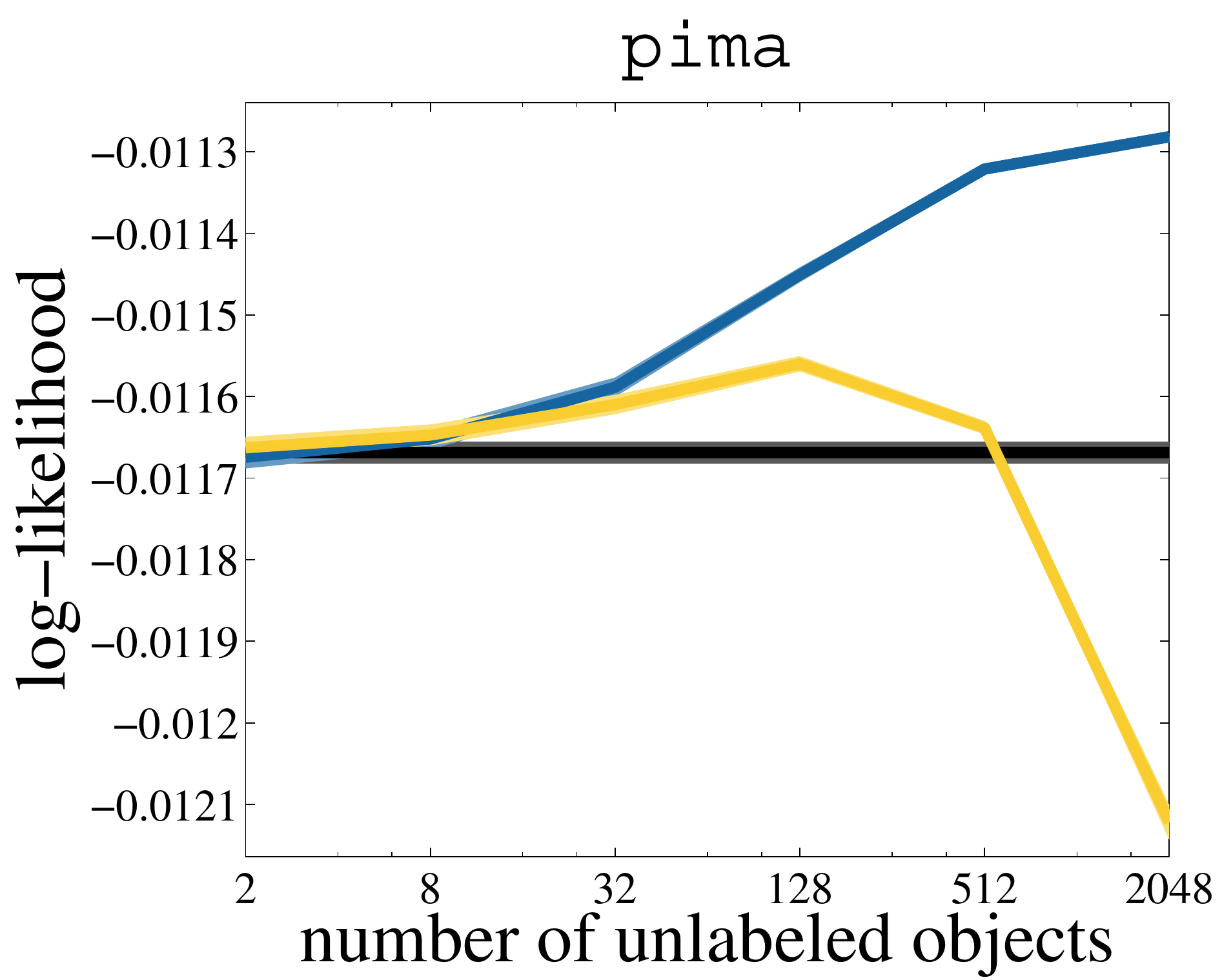} \bigskip \\
\includegraphics[width=0.32\hsize]{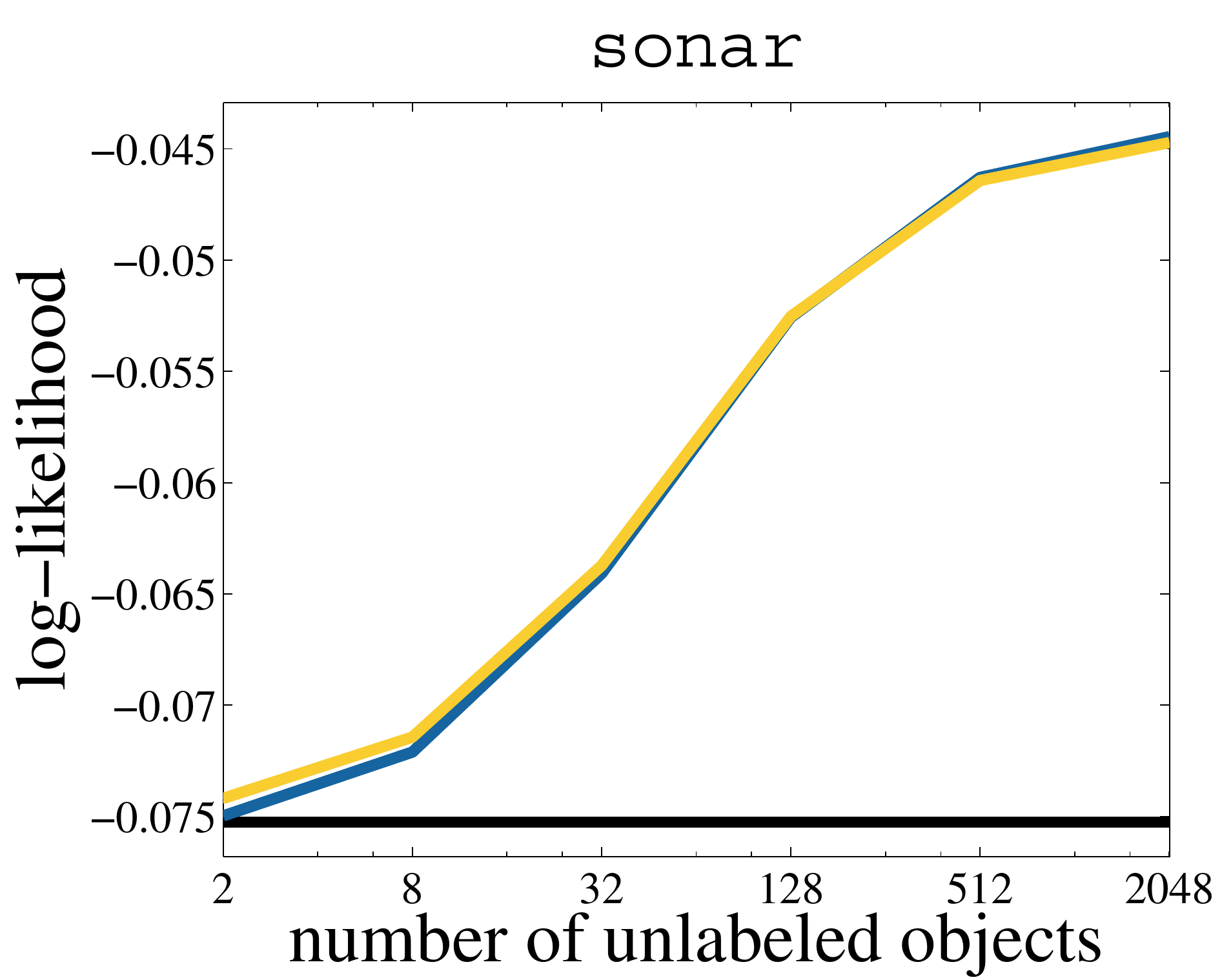}
\includegraphics[width=0.32\hsize]{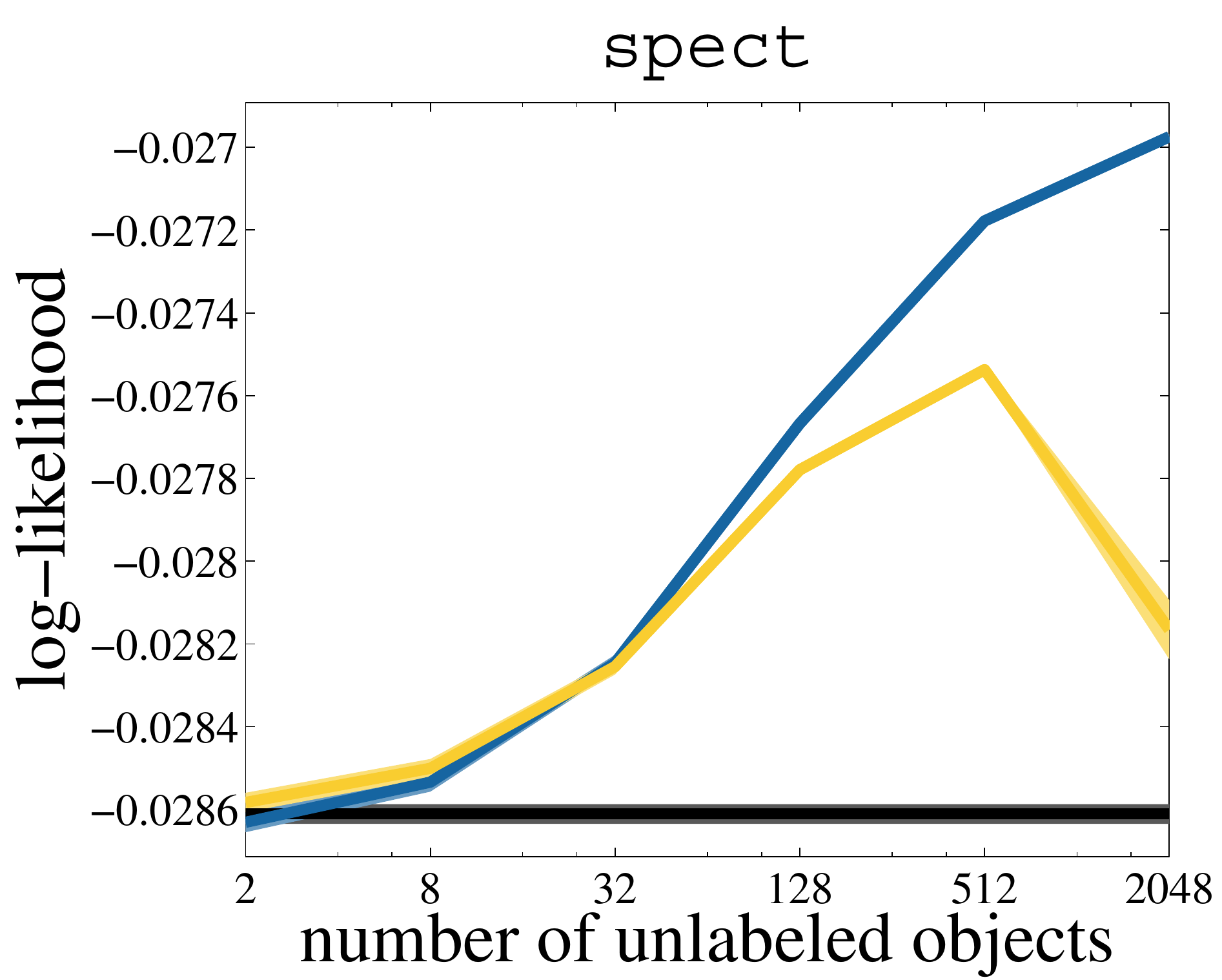} \bigskip \\
\includegraphics[width=0.32\hsize]{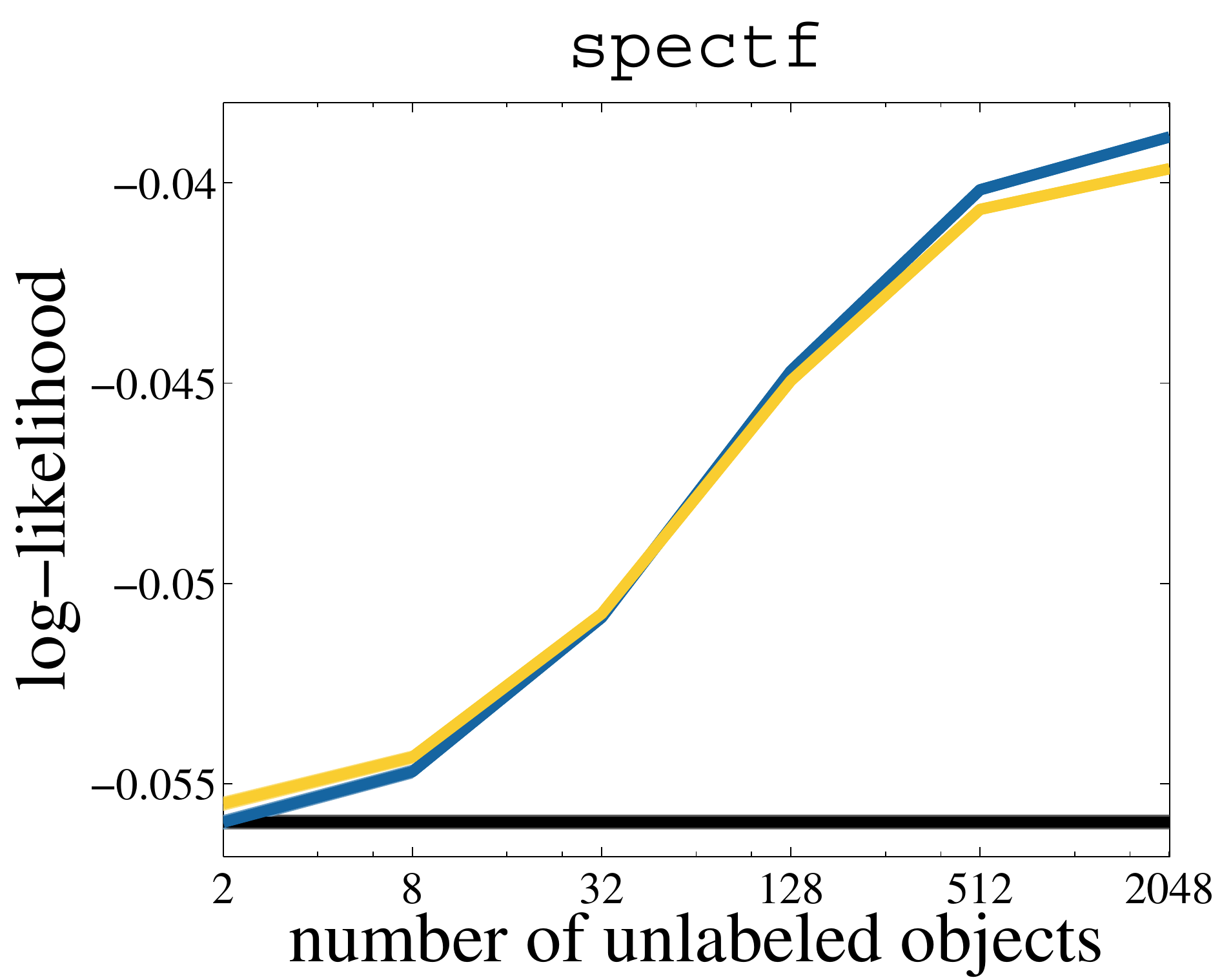}
\includegraphics[width=0.32\hsize]{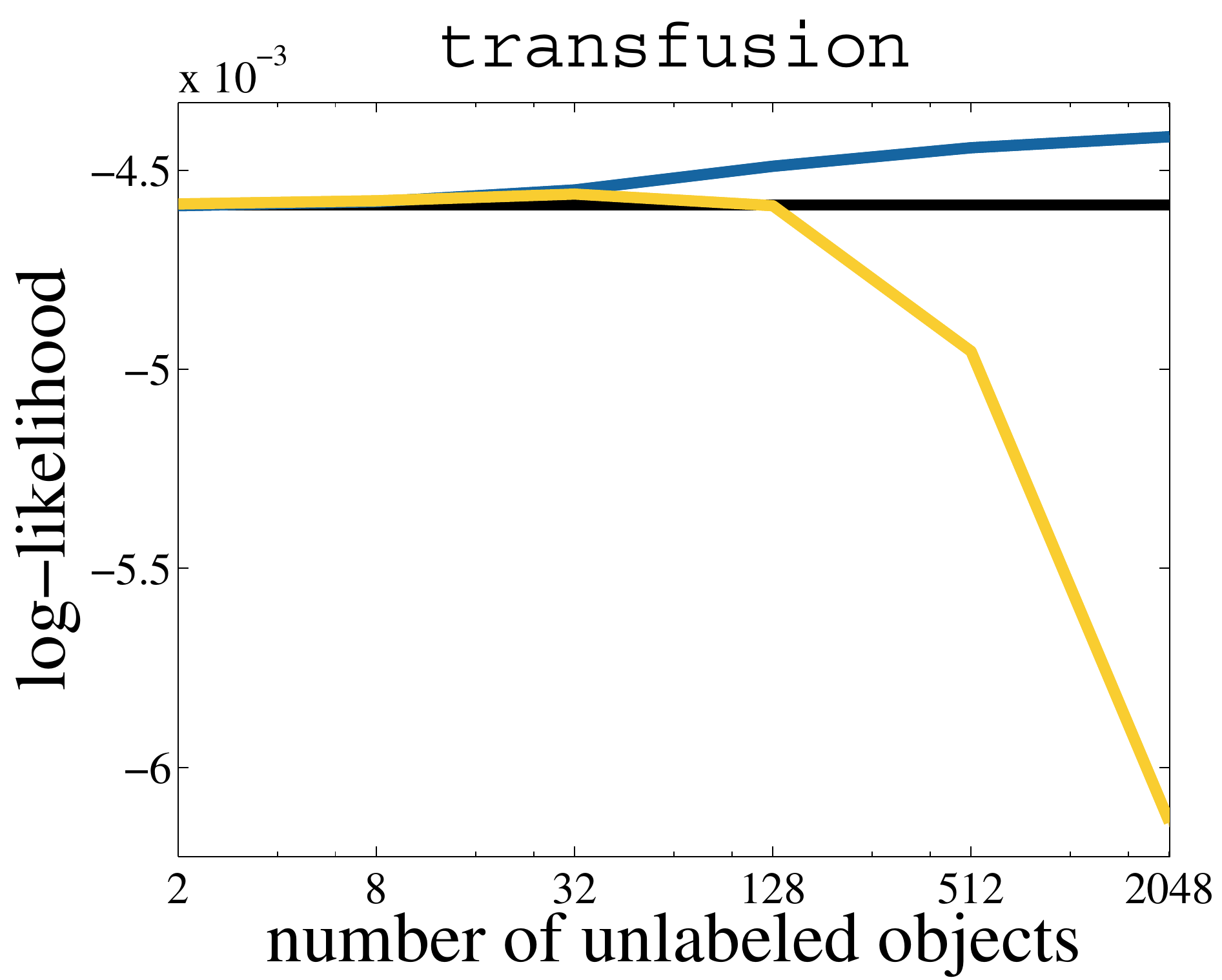}
\includegraphics[width=0.32\hsize]{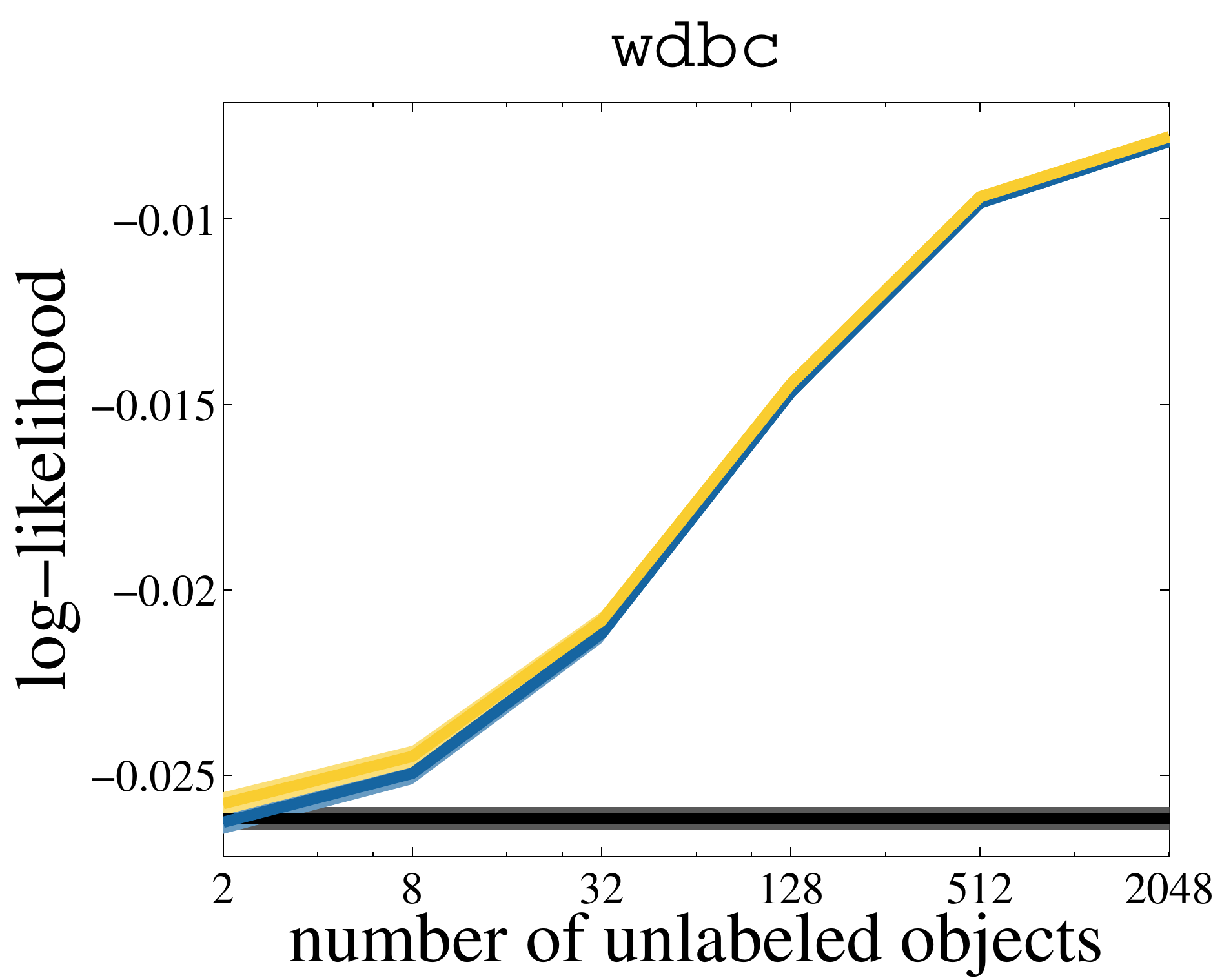}
\caption{The curves for the log-likelihood of the three LDAs corresponding to the error curves in Figure \ref{fig:five}. Note that the blue curve is not always clearly visible as it is more or less fully occluded by the yellow curve.}\label{fig:fiveprime}
\end{figure*}

Following the introductory section, we constructed learning curves both for the expected error rate and the expected log-likelihood (based on the 1000 repetitions).  Figure \ref{fig:one} shows the error rates for the NMCs on the various data sets when only four training samples are available.  Figure \ref{fig:oneprime} shows the error when ten samples are at hand.  The corresponding average log-likelihood curves can be found in Figures \ref{fig:two} and \ref{fig:twoprime}, respectively.  Figure \ref{fig:five} reports the error rates obtained with 100 training samples and using the supervised and semi-supervised LDAs.  Figure \ref{fig:fiveprime} reports on the corresponding log-likelihoods.   The supervised classification performance is displayed in black, self-learners are in yellow (NCS 0580-Y10R), and the constrained versions are in blue (NCS 4055-R95B). The lighter bands around the learning curves give an indication of the standard deviations of the averaged curves, providing an idea of the statistical significance of the differences between the curves.

\section{Discussion and Conclusion}\label{sect:fin}

To start with, it is important to note that when we look at the error rates, behaviors can indeed be quite disperse.  For both classifiers and both constrained and self-learned semi-supervised approaches, there are examples of error rates higher as well as lower than the averaged error rate the regular supervised learners achieve.  Sometimes rather erratic behavior can be noted, like for self-learned NMC on {\tt wdbc} in Figure \ref{fig:one} (yellow curve) and constrained LDA on {\tt haberman} and {\tt transfusion} in Figure \ref{fig:five} (blue curves).  On these last two, also the behavior of self-learned LDA does not seem very regular.  Overall, the performance of the self-learners is very disappointing as only on {\tt wdbc} with 4 labeled training samples, some overall but not very convincing improvements can be observed.  Regarding expected error rates, the constrained approach fares significantly better, showing clear performance improvement in at least 6 of the 16 NMC experiments and in 5 out of 8 of the LDA experiments.  Still, in at least 3 of the 16, classification errors become significantly worse for NMC and, in 5 out of 8 experiments, constrained LDA is not convincing.

Things drastically change indeed when we look at the log-likelihood curves.  For the constrained approaches, looking at Figure \ref{fig:two} and the lower half of Figure \ref{fig:five}, the story is very simple: where for the error rate deteriorations, improvements, and erraticism could be observed, the log-likelihood improves---i.e., increases---in every single case in a smooth, monotonic, and significant way.  Only for LDA on {\tt haberman} and maybe {\tt transfusion}, the constrained approach does not improve as convincingly as in all 22 other cases.

For self-learned NMC and LDA, the results are still mixed.  In many a case, we now do see improvements, but there are still some data sets on which the likelihood decreases.  Notably, for self-learned NMC with 4 labeled samples, the log-likelihood on the test data improves in all cases.  But we do not see the monotonic behavior that the constrained approach displays.  Still, curves are less erratic than those for the error rates.  Nonetheless, it so seems that even if we quantify performance in terms of log-likelihoods, we should be very critical towards self-learning and EM-based approaches.  Behavior definitely is much more regular in terms of its surrogate loss, but performances worse than the supervised approach provides still do occur.

Nevertheless, the results illustrate that it can be interesting to study not only the performance in terms of error rates but also in terms of the surrogate loss.  This is irrespective of the possibility that, ultimately, one might only be interested in the former.  It is encouraging to observe empirically that there seem to be semi-supervised learning schemes that can guarantee improvements in terms of the intrinsic surrogate loss.  This really is a nontrivial observation, as similar guarantees for error rates seem out of the question (unless strict conditions on the data are imposed; cf.\ \cite{castelli95a,ben-david08a,lafferty07a,singh08a}). Although our illustration is in terms of semi-supervised learning, it seems rather plausible that similar observations can be made for other learning settings in which two or more different estimation techniques for the same type of classifier, relying on the same surrogate loss, are compared.  All in all, it is worthwhile considering the behavior of the surrogate in general, as it provides us with a view on a classifier's relative performance that a mere error rate cannot capture.

\bibliographystyle{unsrt}
\bibliography{ws-rv-sample}
\end{document}